\documentclass{article}

\usepackage{PRIMEarxiv}

\usepackage[utf8]{inputenc} 
\usepackage[T1]{fontenc}    
\usepackage{hyperref}       
\usepackage{url}            
\usepackage{booktabs}       
\usepackage{amsfonts}       
\usepackage{nicefrac}       
\usepackage{microtype}      
\usepackage{lipsum}
\usepackage{amsmath}
\usepackage{amssymb}
\usepackage{fancyhdr}       
\usepackage{graphicx}       
\graphicspath{{media/}}     

\title{One2multi Learning
}

\author{
  Heng Tian\textsuperscript{1,*},
  Dequan Gong\textsuperscript{1},
  Tingyi Chen\textsuperscript{1},
  Shaobo Ma\textsuperscript{2,1},
  Su Liu\textsuperscript{3,1},
  Fubo Zhang\textsuperscript{1,*} \\
  \textsuperscript{1}Guangdong Institute of Intelligence
  Science and Technology \\
  \textsuperscript{2}School of Remote Sensing Science and Technology,
  Sun Yat-sen University \\
  \textsuperscript{3}Faculty of Applied Sciences,
  Macao Polytechnic University \\
  \texttt{\{tianheng,gongdequan,chentingyi,zhangfubo\}@gdiist.cn} \\
  \texttt{mashb7@mail2.sysu.edu.cn},
  \texttt{p2412897@mpu.edu.mo} \\
  \textsuperscript{*}Corresponding authors.
}

\begin{document}
\maketitle


\begin{abstract}
Training models to perform visual tasks typically requires task-specific annotations, which become increasingly costly to collect as scene complexity grows. Inspired by humans' ability to acquire visual knowledge from a few single-object examples, we introduce \textbf{one2multi learning}: learning from single-object images with category labels to perform multiple visual tasks in multi-object scenes. To evaluate this capability, we establish a related benchmark comprising an unlabeled single-object base dataset, a few-shot training set of synthetic single-object images with category labels, and multi-object validation and test sets with task-specific annotations. The benchmark covers multi-label image classification, object detection, semantic segmentation, and instance segmentation. To provide a reference for this benchmark, we propose a unified baseline that decouples class-agnostic localization from category recognition. The localization component identifies class-agnostic object regions, while the classification component assigns categories using knowledge acquired from labeled single-object examples. This modular design accommodates new categories by updating category knowledge without retraining the model or modifying the localization component. Together, the benchmark and baseline provide a foundation for future research on one2multi learning. Code and data are available at \url{https://github.com/t1anhen9/one2multi-learning}.
\end{abstract}


\section{Introduction}

In deep learning, a common strategy for improving task performance is to scale up training data, as exemplified by DINOv1--v3 \cite{dinov1,dinov2,dinov3} in vision and GPT-1--3 \cite{gpt-1,gpt-2,gpt-3} in language modeling. Broader data coverage can better align the training distribution with the test distribution, supporting generalization to unseen samples. Conversely, models often struggle with samples outside the training distribution. However, as tasks become more complex, collecting sufficient task-specific data becomes increasingly difficult, particularly in vision.

\begin{figure}[!tbp]
	\centering
	\includegraphics[width=0.7\linewidth]{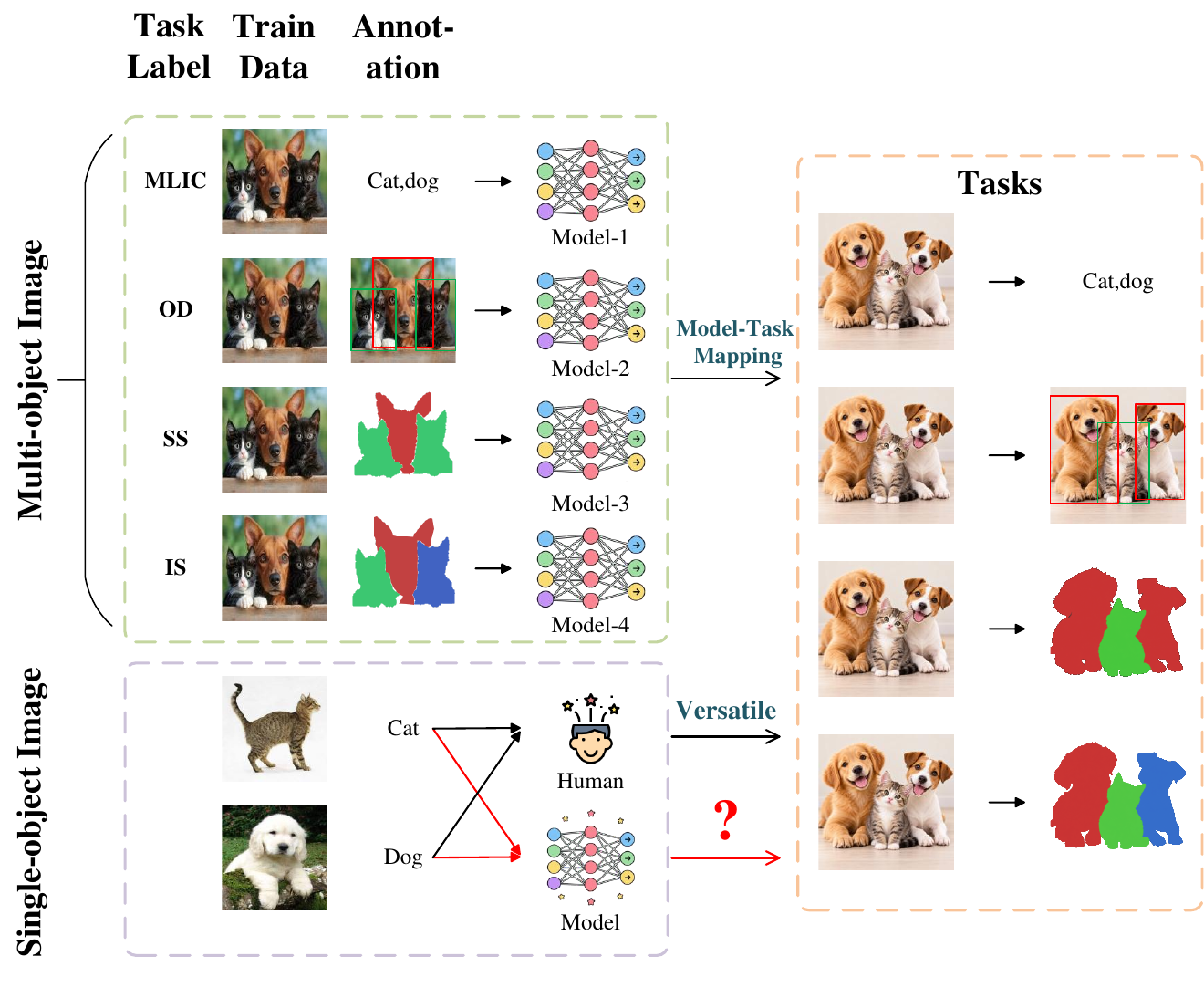}
	\caption{Can models, like humans, learn from a few labeled single-object images to perform multiple visual tasks in multi-object scenes? We introduce one2multi learning to investigate this possibility.}
	\label{fig:framework}
\end{figure}

Visual scenes often contain multiple objects, such as a cat and a dog in the same image. Annotation requirements vary across tasks: multi-label image classification~\cite{mltr} requires identifying all relevant categories, object detection \cite{detr} requires category-labeled bounding boxes, and segmentation \cite{maskdino,segformer} requires pixel-level masks (Figure~\ref{fig:framework}). Training datasets must therefore capture diverse object combinations with task-specific annotations. As scene complexity increases, collecting data that cover these combinations becomes increasingly challenging. Moreover, different tasks typically rely on separately trained models, requiring the appropriate task-specific model at inference time.

\textit{Must training and test data come from the same distribution?} Humans can learn to recognize novel objects from single-object images with category labels, then follow task instructions to identify, localize, or delineate them in multi-object scenes. For example, someone unfamiliar with the okapi~\footnote{A forest mammal native to the Democratic Republic of the Congo, closely related to giraffes and distinguished by zebra-like stripes on its legs.} may recognize and localize it in complex images after viewing only a few labeled examples. Motivated by this ability, we ask whether a model trained on single-object images with only category labels can similarly perform multiple tasks in multi-object scenes (Figure~\ref{fig:framework}).

Inspired by humans' ability to learn from single-object examples and recognize objects in complex scenes, we introduce \emph{one2multi learning}, a problem setting that studies whether models can learn from single-object images and perform multiple visual tasks in multi-object scenes. To evaluate this capability, we establish a benchmark comprising an unlabeled base dataset, a few-shot training set with category labels, and validation and test sets with task-specific annotations. Both the base dataset and the training set contain single-object images, whereas the validation and test sets contain multi-object scenes. The few-shot training set reflects learning novel categories from limited labeled examples. To reduce data collection and annotation costs, we generate single-object images using a text-to-image model \cite{flux2klein9b} guided by prompts produced by a large language model \cite{deepseekv41flash}. Evaluation covers four tasks: multi-label image classification~\cite{mltr,mldecoder}, object detection~\cite{detr,fastrcnn}, semantic segmentation~\cite{segformer,fcn}, and instance segmentation~\cite{maskdino,maskrcnn}.

One2multi learning offers a unified supervision setting for multiple visual tasks: the same category-labeled single-object examples support tasks that conventionally require different training annotations. This reduces the need to collect separate box- or mask-annotated training sets for each task. Single-object examples also allow direct control over the number of training images per category and can be generated using text-to-image models~\cite{flux2klein9b}, facilitating the construction of balanced training sets. Together, these properties make one2multi learning a promising direction for multi-task visual understanding with limited annotation effort.

\begin{figure}[t]
	\centering
	\includegraphics[width=0.7\linewidth]{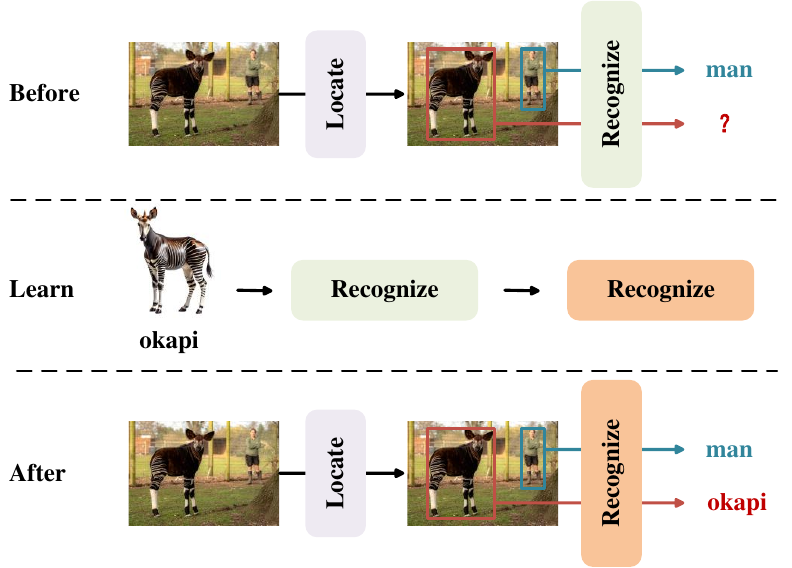}
	\caption{
		Motivation for decoupling localization from recognition.
		Before learning, the unfamiliar okapi is localized
		but its category is unknown. After learning from
		labeled single-object examples, it can be recognized
		while the localization component remains unchanged.
	}
	\label{fig:intuition}
\end{figure}

We also introduce a baseline for one2multi learning to provide a reference for future research. Figure~\ref{fig:intuition} illustrates its motivation: an unfamiliar okapi can be localized before its category is known, and learning from labeled single-object examples enables its recognition without changing the localization component. Motivated by this distinction between locating objects and identifying their categories~\cite{rfcn,glip}, the baseline decouples class-agnostic localization from category recognition. The localization component identifies candidate object regions, while the recognition component uses category knowledge acquired from labeled single-object images to classify images and regions. This design supports multi-label image classification, object detection, semantic segmentation, and instance segmentation within a unified framework. New categories can be incorporated by updating category knowledge without retraining the model or modifying the localization component. This process mirrors how humans learn to recognize new objects from a few labeled examples and aligns with the motivation of one2multi learning.
Our contributions are threefold:
\begin{itemize}
	\item We introduce one2multi learning, a problem setting in which models learn from a small number of labeled single-object images to perform multiple visual tasks in multi-object scenes.
	\item We establish a benchmark for one2multi learning covering multi-label image classification, object detection, semantic segmentation, and instance segmentation.
	\item We develop a unified baseline that decouples class-agnostic localization from category recognition, enabling new categories to be incorporated by updating category knowledge without retraining the model or modifying the localization component.
\end{itemize}

\section{Related Work}
\noindent\textbf{Supervised Learning.}
Conventional supervised approaches~\cite{resnet,maskrcnn} typically rely on task-specific training annotations. Object detectors such as DETR \cite{detr}, for example, are trained using category labels and bounding boxes. Their generalization is commonly studied under the assumption that training and test samples are independently drawn from the same distribution, although practical deployment often involves distribution shifts. Collecting task-specific annotations is costly, particularly for multi-object scenes requiring boxes or masks for individual objects. To reduce this dependence, one2multi learning investigates whether models can learn category knowledge from a few labeled single-object images and perform multiple tasks in multi-object scenes without box or mask supervision during training. 

\noindent\textbf{Weakly Supervised Learning.}
Weakly supervised learning~\cite{weaklySL} uses incomplete, imprecise, or noisy annotations to reduce supervision requirements. For example, multi-label learning from single positive labels~\cite{singlepositive} provides one observed positive label per training image, leaving all other category labels unknown, although the image may contain multiple objects. In contrast, one2multi learning constrains the training images themselves to contain a single target object and uses a few category-labeled examples to support classification, detection, and segmentation in multi-object scenes. Rather than reducing annotation requirements for a single target task, this setting investigates whether the same single-object learning material can support multiple visual tasks with different output requirements.

\noindent\textbf{Unsupervised Learning.}
Unsupervised visual learning includes self-supervised representation learning~\cite{dinov1,mae} and class-agnostic object discovery~\cite{lost,cltler}, which localizes objects without identifying their categories. In contrast, one2multi learning requires category recognition alongside localization. Our baseline builds on MaskCut~\cite{cltler} and introduces an adaptive variant that determines the number of candidate masks per image, stopping extraction when the remaining uncovered region or a newly extracted region falls below a minimum size.

\noindent\textbf{Few-Shot Learning.}
Inspired by humans' ability to learn new categories from a few examples, one2multi learning provides only a small number of labeled single-object images per category. Unlike common episodic few-shot protocols organized into support and query sets, our benchmark uses fixed training, validation, and test splits with a controlled number of training examples per category. In multi-object tasks, low-label regimes may instead use a fraction of the available annotations, such as 5\%~\cite{cltler}. Sampling multi-object images does not necessarily yield balanced category coverage, since categories co-occur and instance counts vary across images. By using single-object training images, our benchmark enables direct control over the number of labeled examples per category.

\noindent\textbf{Transfer Learning.}
Transfer learning is widely used in computer vision~\cite{decaf,transferable} and natural language processing~\cite{bert,t5}, often through pretraining followed by task-specific fine-tuning. Neither unlabeled pretraining nor distribution differences alone distinguish one2multi learning from transfer learning. Instead, one2multi learning defines a specific setting in which models learn from single-object images with limited category-level supervision and perform multiple tasks in multi-object scenes. Its defining constraint concerns the learning material and available annotations, rather than a particular pretraining or fine-tuning strategy.

\noindent\textbf{Foundation Models.}
Visual foundation models such as SAM~\cite{sam,sam2,sam3} and Florence-2~\cite{florence2} use large-scale training data to support promptable segmentation or multiple vision tasks. Despite their versatility, large models can impose substantial computational and memory costs on resource-constrained devices. One2multi learning explores a complementary setting based on a few category-labeled single-object images. Our baseline decouples class-agnostic localization from category recognition, allowing new categories to be incorporated by updating category knowledge while keeping the localization component unchanged.

\section{One2multi Learning}
\label{sec:p0e6a3c7ad280ed4a294a6cf8be8175b4}

\subsection{Problem Definition}
\label{sec:p7863e7915170b0e12f1a24a377b81a20}
One2multi learning studies how a few category-labeled single-object examples support multi-object scene understanding. ``One'' denotes a single target object per learning example, labeled only by category, without additional task-specific annotations such as bounding boxes or masks. ``Multi'' refers to evaluation scenes that may contain multiple objects, with task-specific annotations. Let $\mathcal{C}$ denote a set of target categories. The training set is defined as \( \mathfrak{D}_o = \{(\mathbf{x}_i,c_i)\}_{i=1}^{M}, c_i \in \mathcal{C} \), where each image $\mathbf{x}_i$ contains a single target object annotated only with its category label $c_i$. In the few-shot setting, only $K$ labeled examples are provided for each category. After learning from $\mathfrak{D}_o$, the model $\Psi$ takes a multi-object image $\mathbf{x}$ and a task specification $t$ as inputs and produces \( \widehat{\mathbf{y}}_t = \Psi(\mathbf{x},t) \), where $\widehat{\mathbf{y}}_t$ denotes the prediction for task $t$. Training and evaluation share the same target categories but differ in object composition, scene complexity, and task requirements.

\subsection{Benchmark Construction and Evaluation}
\label{sec:paca1c0c0e6399d1a0ba69b42e744e62c}

We instantiate one2multi learning through a benchmark that combines existing datasets with synthetic single-object images. It covers four tasks: multi-label image classification (MLIC), object detection (OD), semantic segmentation (SS), and instance segmentation (IS). The benchmark comprises an unlabeled single-object base set, category-labeled single-object training sets, and task-specific multi-object validation and test sets for hyperparameter selection and final evaluation. Further construction and evaluation details are provided in Section~\ref{sec:supp_one2multi} of the supplementary material.

\subsubsection{Datasets}
\label{sec:p53709268d3ec7cfdbbb3280d574d2fbe}

\noindent\textbf{Base Set $\mathfrak{D}_b$.}
We use ImageNet-1K~\cite{imagenet} with category labels discarded for self-supervised representation learning. Its object-centric images approximate the intended single-object learning material, although some images may contain multiple objects.

\begin{figure}[t]
	\centering
	\includegraphics[width=\linewidth]{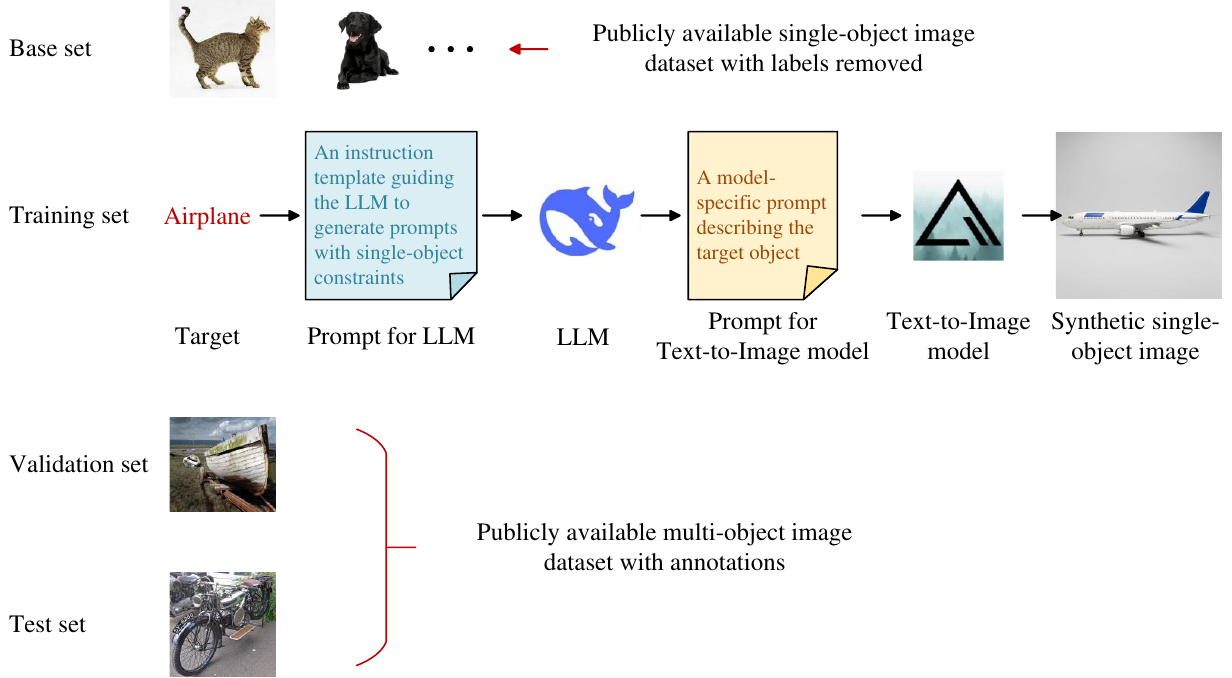}
	\caption{
		Benchmark construction.
		A language model converts target categories and single-object
		constraints into prompts for a text-to-image model,
		producing synthetic training images with category labels.
		The base, validation, and test sets use existing datasets.
	}
	\label{fig:41182dc8}
\end{figure}

\noindent\textbf{Training Sets $\mathfrak{D}_o$.}
We generate single-object training images using a language model~\cite{deepseekv41flash} and a text-to-image model~\cite{flux2klein9b}, as illustrated in Figure~\ref{fig:41182dc8}. Given target categories, sample counts, and background conditions, the language model produces prompts that guide image generation. For each category, we generate 10 images with backgrounds and 10 without backgrounds. Learning uses only category labels, without bounding boxes, masks, or other task-specific annotations. Table~\ref{tab:prompt_pipeline} in the supplementary material provides example LLM inputs, generated prompts, and corresponding images.

\noindent\textbf{Validation and Test Sets.}
We use established datasets for multi-object evaluation: VOC~\cite{voc}, COCO~\cite{coco}, NUS-WIDE~\cite{nus}, and VG500~\cite{vg500} for MLIC, and VOC and COCO for OD, SS, and IS. Each task uses its corresponding annotations and data splits, detailed in Section~\ref{sec:supp_o2m_datasets}.

\subsubsection{Evaluation Protocol}
\label{sec:p8de4351e09e10d4a1298f30d06b72b2f}
For MLIC, the primary metric is image-averaged label-set IoU. For OD, we report Box $\mathrm{AP}_{50}$ on VOC and Box $\mathrm{AP}_{50:95}$ on COCO. For SS, we report mIoU, including background and excluding ignored pixels. For IS, we report COCO-style Mask $\mathrm{AP}_{50:95}$ on both datasets. Additional metrics and evaluation details are provided in Section~\ref{sec:supp_o2m_protocol}.

\section{Baseline for One2multi Learning}
\label{sec:pcc914eda9010c895f25245151653b64c}

Alongside the benchmark, we introduce a baseline that decouples object localization from category recognition. This design is motivated by the distinction between perceiving objects and identifying their categories through acquired knowledge~\cite{learnlikepeople}. We instantiate this idea by combining class-agnostic region proposals with category prototypes: new categories are incorporated by constructing additional prototypes, while the visual encoder and localization component remain unchanged. While R-FCN~\cite{rfcn} and GLIP~\cite{glip} integrate localization and recognition within a single model, our formulation allows the two components to be implemented as separate models. This baseline provides a reference for future research on one2multi learning.

\begin{figure*}[t]
	\centering
	\includegraphics[width=0.9\linewidth]{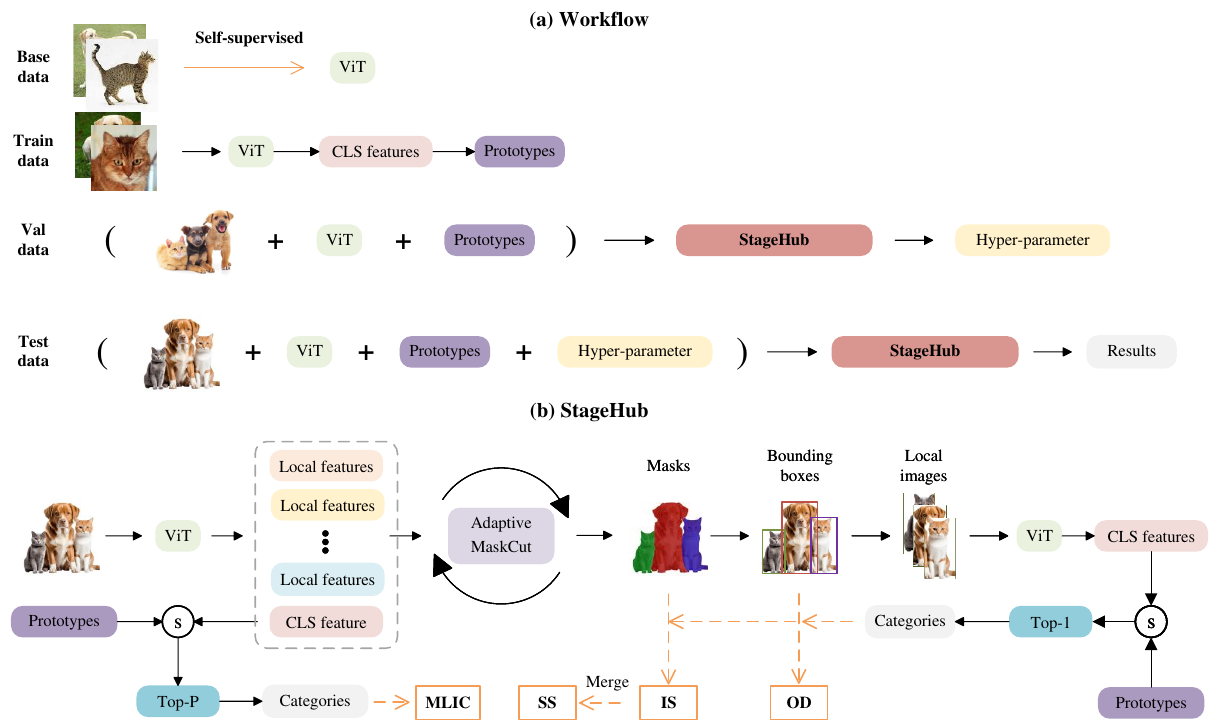}
	\caption{
		Overview of the baseline. (a) Self-supervised learning on the base set provides a visual encoder, and single-object training images are used to construct category prototypes. Hyperparameters are selected on the validation set and fixed for test evaluation. (b) Image-level prototype matching supports MLIC, while AMC generates candidate masks and boxes for region-level recognition. The classified boxes and masks support OD and IS, respectively, and merging the classified masks produces SS predictions.
	}
	\label{fig:464b048f}
\end{figure*}

Figure~\ref{fig:464b048f} summarizes the baseline workflow. Self-supervised learning on the unlabeled base set provides a visual encoder. Labeled single-object training images are used to construct category prototypes. StageHub reuses image-level recognition, class-agnostic localization, and region-level recognition outputs to support the four target tasks. Validation data guide hyperparameter selection, and the resulting configuration is fixed for test evaluation. Further implementation details are provided in Section~\ref{sec:supp_baseline} of the supplementary material.

\subsection{Self-Supervised Learning}
\label{sec:pde497933f3dc5a05e268e7e05f2774f4}
We first pretrain a Vision Transformer (ViT)~\cite{vit} on the $\mathfrak{D}_b$ through self-supervised learning~\cite{dinov1,mae,simclr}. The encoder is then frozen and shared across tasks, providing global features for category recognition and local features for class-agnostic localization. Publicly available self-supervised checkpoints can also be used to instantiate this stage without additional pretraining.

\subsection{Prototypes}
\label{sec:Prototypes}

We represent each category by a prototype constructed from \( \mathfrak{D}_{o} \).
Given an image $\mathbf{x} \in \mathfrak{D}_{o}$, the frozen ViT extracts
\begin{equation}
	\mathbf{F} = \operatorname{ViT}(\mathbf{x})
	\in \mathbb{R}^{(N+1)\times D},
	\label{eq:26a62415}
\end{equation}
where $\mathbf{F}[0,:]$ is the global CLS feature, $N$ is the number of patch tokens, and $D$ is the feature dimension. Let $\mathfrak{D}_{o,c}$ denote the subset of single-object training images belonging to category $c$. We average their CLS features and apply $L_2$ normalization to obtain the category prototype:
\begin{equation}
	\boldsymbol{\mu}_c =
	\frac{1}{|\mathfrak{D}_{o,c}|}
	\sum_{\mathbf{x}\in\mathfrak{D}_{o,c}}
	\mathbf{F}[0,:],
	\qquad
	\mathbf{p}_c =
	\frac{\boldsymbol{\mu}_c}{\|\boldsymbol{\mu}_c\|_2}.
	\label{eq:class_prototype}
\end{equation}
The prototypes are shared across tasks and serve as category references for both image-level and region-level recognition.

\subsection{StageHub}
\label{sec}

StageHub organizes the outputs of different processing stages into predictions for multiple visual tasks. Its central idea is to reuse stage-wise results, including image-level category scores, candidate masks, bounding boxes, and region labels, rather than build a separate prediction pipeline for each task. Image-level category matching provides MLIC predictions. Along the region-level pathway, localization produces candidate masks and boxes, and prototype matching assigns their categories. The labeled boxes and masks yield OD and IS predictions, respectively, while merging the classified masks produces an SS prediction.

\noindent\textbf{Image-Level Recognition.}
For MLIC, we compute cosine similarities between the normalized global CLS feature and class prototypes. We convert these similarities into probabilities using a temperature-scaled softmax and select labels through top-$p$ selection~\cite{top-p}. Specifically, we rank categories by probability and retain the smallest set of top-ranked categories whose cumulative probability reaches or exceeds $p$.

\noindent\textbf{Adaptive MaskCut.}
For object localization, we extend MaskCut~\cite{cltler} to adapt the number of candidate regions to each image. We construct an affinity graph from normalized local features and iteratively extract class-agnostic regions using normalized cuts~\cite{normalizedcuts}. Features at previously covered locations are suppressed to reduce duplicate proposals. At each iteration, we retain the largest connected component of the newly uncovered region. Extraction stops when this component or the remaining uncovered area falls below a minimum size, or when no valid partition is found. The resulting patch-level masks are upsampled to the original image resolution and refined with DenseCRF~\cite{densecrf}. We retain non-empty refined masks that satisfy an IoU consistency criterion with their coarse counterparts.

\noindent\textbf{Bounding Box Generation.}
For each retained mask, we compute the tight axis-aligned bounding box enclosing its foreground pixels. These boxes provide localization predictions and define the crops used for region recognition.

\noindent\textbf{Region Recognition.}
We crop each candidate region using its bounding box and set pixels outside its mask to black. The frozen ViT extracts the crop's final-layer CLS feature, which is $L_2$-normalized and compared with the class prototypes. We assign the category with the highest cosine similarity and use that similarity as the prediction score. Both the category and score are attached to the corresponding box and mask. 

\noindent\textbf{Task Readout.}
Each task reads out the relevant stage outputs: MLIC uses the selected image-level labels, OD uses the classified boxes, and IS uses the classified instance masks. For SS, we merge these masks into a semantic label map, assigning overlapping pixels to the highest-scoring mask and uncovered pixels to background. This organization reuses intermediate results across tasks without introducing separately trained task-specific prediction heads.

By decoupling localization from recognition, our baseline incorporates new categories through additional prototypes constructed from labeled single-object images, without retraining the encoder or localization module. This design aligns with how humans acquire new category knowledge while reusing existing localization capabilities.

\begin{table*}[!tbp]
	\centering
	\caption{
		Multi-label image classification results measured by
		image-averaged label-set IoU.
		Bold indicates the best result in each column.
	}
	\label{tab:mlic_all}
	\resizebox{\textwidth}{!}{%
		\begin{tabular}{ll*{4}{ccc}}
			\toprule
			                                                                                          &     &   \multicolumn{3}{c}{VOC\textsubscript{mlic}}    &   \multicolumn{3}{c}{COCO\textsubscript{mlic}}   & \multicolumn{3}{c}{NUS-WIDE\textsubscript{mlic}} & \multicolumn{3}{c}{VG500\textsubscript{mlic}} \\ \cmidrule
			(lr){3-5}
			\cmidrule(lr){6-8}
			\cmidrule(lr){9-11}
			\cmidrule(lr){12-14}
			Encoder & BG  &     1-shot     &     5-shot     &    10-shot     &     1-shot     &     5-shot     &    10-shot     &     1-shot     &     5-shot     &    10-shot     &    1-shot     &    5-shot     &    10-shot    \\ \midrule
			DINO ViT-S/8                                                                              & Yes &     51.64      &     63.28      &     63.54      &     32.73      & \textbf{37.03} & \textbf{38.70} &     14.71      &     17.51      &     17.30      &     5.89      &     6.84      &     6.76      \\
			DINO ViT-S/8                                                                              & No  &     56.71      &     64.59      &     66.06      &     33.48      &     36.00      &     36.93      &     13.20      &     18.70      &     19.39      &     5.94      &     6.83      &     6.84      \\
			DINO ViT-S/16                                                                             & Yes &     47.27      &     61.20      &     61.42      &     28.56      &     33.32      &     35.69      &     14.03      &     16.97      &     16.92      &     5.61      &     6.73      &     6.76      \\
			DINO ViT-S/16                                                                             & No  &     55.35      &     64.68      &     65.53      &     31.14      &     34.06      &     35.34      & \textbf{14.72} & \textbf{19.59} & \textbf{20.28} &     5.77      &     6.55      &     6.69      \\
			DINO ViT-B/8                                                                              & Yes &     49.58      &     63.38      &     63.87      &     32.79      &     36.98      &     38.64      &     13.94      &     17.08      &     17.17      &     6.02      & \textbf{7.05} & \textbf{7.11} \\
			DINO ViT-B/8                                                                              & No  & \textbf{57.02} &     65.39      &     66.02      & \textbf{33.61} &     36.04      &     37.51      &     13.14      &     18.09      &     18.91      & \textbf{6.22} &     7.01      &     7.00      \\
			DINO ViT-B/16                                                                             & Yes &     49.18      &     61.18      &     62.16      &     30.26      &     35.06      &     36.58      &     13.23      &     16.49      &     16.35      &     5.82      &     6.91      &     6.93      \\
			DINO ViT-B/16                                                                             & No  &     56.24      & \textbf{65.70} & \textbf{66.50} &     33.27      &     35.62      &     36.96      &     14.17      &     19.50      &     20.12      &     6.09      &     6.88      &     6.93      \\ \midrule
			MAE ViT-B/16                                                                              & Yes &     14.08      &     25.86      &     25.74      &      6.39      &      9.69      &     11.81      &      4.47      &      6.42      &      7.20      &     3.17      &     4.16      &     4.49      \\
			MAE ViT-B/16                                                                              & No  &     17.23      &     25.77      &     26.19      &      7.78      &      8.45      &     10.02      &      3.51      &      8.48      &      8.51      &     3.56      &     4.33      &     4.40      \\
			MAE ViT-L/16                                                                              & Yes &     12.84      &     18.81      &     21.14      &      5.34      &      7.99      &      9.93      &      3.15      &      4.36      &      5.03      &     3.00      &     3.22      &     3.42      \\
			MAE ViT-L/16                                                                              & No  &     10.12      &     24.63      &     23.22      &      5.68      &      7.81      &      8.15      &      2.88      &      7.28      &      4.99      &     2.80      &     3.49      &     3.48      \\
			MAE ViT-H/14                                                                              & Yes &     10.72      &     13.23      &     14.44      &      4.06      &      6.92      &      7.87      &      3.18      &      3.87      &      4.21      &     2.79      &     2.93      &     3.02      \\
			MAE ViT-H/14                                                                              & No  &      7.42      &     12.80      &     14.57      &      4.31      &      5.68      &      5.81      &      2.60      &      6.10      &      4.37      &     2.68      &     3.20      &     3.15      \\ \bottomrule
		\end{tabular}%
	}
\end{table*}

\begin{table}[t]
	\centering
	\caption{
		Object detection results are reported as Box \(\mathrm{AP}_{50}\) on VOC\textsubscript{od} and Box \(\mathrm{AP}_{50:95}\) on COCO\textsubscript{od}.
	}
	\label{tab:od_all}
		\begin{tabular}{lllcccccc}
			\toprule
			                                           &     &        & \multicolumn{3}{c}{VOC\textsubscript{od}: Box $\mathrm{AP}_{50}$} & \multicolumn{3}{c}{COCO\textsubscript{od}: Box $\mathrm{AP}_{50:95}$} \\ \cmidrule
			(lr){4-6}
			\cmidrule(lr){7-9}
			Encoder & BG  & Method &     1-shot     &     5-shot     &             10-shot             &    1-shot     &    5-shot     &                10-shot                \\ \midrule
			DINO ViT-S/8                               & Yes & SSP    &      7.06      &     10.16      &              10.20              &     1.18      &     1.48      &                 1.55                  \\
			                                           &     & AMC    &     23.67      &     30.88      &              31.15              &     5.50      &     6.21      &                 6.49                  \\
			DINO ViT-S/8                               & No  & SSP    &      7.28      &     10.15      &              10.27              &     1.57      &     1.84      &                 1.88                  \\
			                                           &     & AMC    &     26.72      &     31.78      &              32.06              &     6.00      & \textbf{6.87} &             \textbf{7.05}             \\ \midrule
			DINO ViT-S/16                              & Yes & SSP    &      5.76      &      9.43      &              9.59               &     1.01      &     1.25      &                 1.30                  \\
			                                           &     & AMC    &     19.02      &     26.67      &              27.25              &     4.15      &     4.51      &                 4.62                  \\
			DINO ViT-S/16                              & No  & SSP    &      6.77      &      9.64      &              9.52               &     1.36      &     1.64      &                 1.69                  \\
			                                           &     & AMC    &     23.28      &     28.03      &              28.16              &     4.60      &     5.27      &                 5.44                  \\ \midrule
			DINO ViT-B/8                               & Yes & SSP    &      6.25      &      9.06      &              9.24               &     1.13      &     1.41      &                 1.43                  \\
			                                           &     & AMC    &     21.91      &     30.34      &              31.90              &     5.18      &     5.94      &                 6.10                  \\
			DINO ViT-B/8                               & No  & SSP    &      6.48      &      8.69      &              8.57               &     1.40      &     1.62      &                 1.65                  \\
			                                           &     & AMC    & \textbf{29.16} & \textbf{34.12} &         \textbf{34.39}          & \textbf{6.02} &     6.76      &                 6.97                  \\ \midrule
			DINO ViT-B/16                              & Yes & SSP    &      6.80      &     10.11      &              10.18              &     1.12      &     1.42      &                 1.47                  \\
			                                           &     & AMC    &     21.26      &     28.31      &              29.28              &     4.31      &     4.87      &                 4.98                  \\
			DINO ViT-B/16                              & No  & SSP    &      7.32      &     10.02      &              10.04              &     1.57      &     1.79      &                 1.86                  \\
			                                           &     & AMC    &     26.52      &     30.83      &              31.12              &     5.00      &     5.70      &                 5.86                  \\ \midrule
			MAE ViT-B/16                               & Yes & SSP    &      0.80      &      1.22      &              1.57               &     0.14      &     0.15      &                 0.18                  \\
			                                           &     & AMC    &      0.77      &      3.27      &              2.57               &     0.32      &     0.18      &                 0.32                  \\
			MAE ViT-B/16                               & No  & SSP    &      2.15      &      4.12      &              3.68               &     0.61      &     0.63      &                 0.60                  \\
			                                           &     & AMC    &      3.82      &      5.40      &              5.83               &     0.73      &     0.92      &                 0.80                  \\ \midrule
			MAE ViT-L/16                               & Yes & SSP    &      0.76      &      1.64      &              1.80               &     0.12      &     0.12      &                 0.17                  \\
			                                           &     & AMC    &      0.52      &      0.55      &              0.99               &     0.03      &     0.05      &                 0.07                  \\
			MAE ViT-L/16                               & No  & SSP    &      1.26      &      2.77      &              2.95               &     0.43      &     0.45      &                 0.45                  \\
			                                           &     & AMC    &      0.49      &      1.15      &              1.09               &     0.15      &     0.16      &                 0.16                  \\ \midrule
			MAE ViT-H/14                               & Yes & SSP    &      0.53      &      0.63      &              0.68               &     0.08      &     0.07      &                 0.11                  \\
			                                           &     & AMC    &      0.23      &      0.19      &              0.23               &     0.01      &     0.03      &                 0.06                  \\
			MAE ViT-H/14                               & No  & SSP    &      0.50      &      1.55      &              1.60               &     0.23      &     0.28      &                 0.27                  \\
			                                           &     & AMC    &      0.31      &      0.62      &              0.71               &     0.07      &     0.09      &                 0.08                  \\ \bottomrule
		\end{tabular}%
\end{table}

\begin{table}[!t]
	\centering
	\caption{
		Semantic segmentation results measured by mIoU,
		including background and excluding ignored pixels.
	}
	\label{tab:ss_all}
	\begin{tabular}{llcccccc}
		\toprule
		                                         &     &            \multicolumn{3}{c}{VOC\textsubscript{ss}}            &           \multicolumn{3}{c}{COCO\textsubscript{ss}}            \\ \cmidrule
		(lr){3-5}
		\cmidrule(lr){6-8}
		Encoder & BG  &     1-shot     &     5-shot     &    10-shot     &     1-shot     &     5-shot     &    10-shot     \\ \midrule
		DINO ViT-S/8                                 & Yes &     36.61      & \textbf{46.44} &     46.41      &     20.00      &     22.65      &     23.53      \\
		DINO ViT-S/8                                 & No  & \textbf{41.59} &     43.07      &     45.27      & \textbf{20.42} & \textbf{23.30} & \textbf{24.43} \\
		DINO ViT-S/16                                & Yes &     35.03      &     41.76      &     42.50      &     15.99      &     18.15      &     19.11      \\
		DINO ViT-S/16                                & No  &     37.30      &     40.57      &     41.91      &     17.99      &     20.24      &     21.06      \\
		DINO ViT-B/8                                 & Yes &     30.29      &     44.26      &     42.95      &     19.22      &     21.07      &     22.14      \\
		DINO ViT-B/8                                 & No  &     38.20      &     45.16      & \textbf{46.52} &     19.98      &     21.94      &     23.31      \\
		DINO ViT-B/16                                & Yes &     29.80      &     39.04      &     37.21      &     16.53      &     18.76      &     20.01      \\
		DINO ViT-B/16                                & No  &     39.09      &     37.73      &     39.34      &     18.91      &     20.64      &     21.81      \\ \midrule
		MAE ViT-B/16                                 & Yes &      5.08      &     15.95      &     12.38      &      2.59      &      2.43      &      3.56      \\
		MAE ViT-B/16                                 & No  &      9.24      &     13.93      &     13.79      &      4.12      &      5.52      &      5.06      \\
		MAE ViT-L/16                                 & Yes &      4.49      &      4.58      &      7.09      &      1.19      &      1.64      &      2.39      \\
		MAE ViT-L/16                                 & No  &      3.00      &      6.92      &      5.23      &      1.86      &      2.62      &      2.43      \\
		MAE ViT-H/14                                 & Yes &      3.76      &      4.34      &      4.80      &      0.98      &      0.99      &      1.50      \\
		MAE ViT-H/14                                 & No  &      3.54      &      3.77      &      4.19      &      1.28      &      1.57      &      1.33      \\ \bottomrule
	\end{tabular}
\end{table}

\begin{table}[!t]
	\centering
	\caption{
		Instance segmentation results measured by
		Mask $\mathrm{AP}_{50:95}$.
	}
	\label{tab:is_all}
	\begin{tabular}{llcccccc}
		\toprule
		                                         &     &    \multicolumn{3}{c}{VOC\textsubscript{is}}     &  \multicolumn{3}{c}{COCO\textsubscript{is}}   \\ \cmidrule
		(lr){3-5}
		\cmidrule(lr){6-8}
		Encoder & BG  &     1-shot     &     5-shot     &    10-shot     &    1-shot     &    5-shot     &    10-shot    \\ \midrule
		DINO ViT-S/8                             & Yes &     16.96      & \textbf{21.00} & \textbf{20.89} &     4.99      &     5.73      &     6.00      \\
		DINO ViT-S/8                             & No  & \textbf{18.58} &     20.38      &     20.69      & \textbf{5.40} & \textbf{6.23} & \textbf{6.39} \\
		DINO ViT-S/16                            & Yes &     12.84      &     15.89      &     16.48      &     3.76      &     4.08      &     4.21      \\
		DINO ViT-S/16                            & No  &     13.74      &     16.28      &     16.57      &     4.17      &     4.80      &     4.95      \\
		DINO ViT-B/8                             & Yes &     14.38      &     17.94      &     18.18      &     4.60      &     5.33      &     5.49      \\
		DINO ViT-B/8                             & No  &     16.73      &     18.13      &     19.00      &     5.34      &     5.97      &     6.15      \\
		DINO ViT-B/16                            & Yes &     12.90      &     15.60      &     15.57      &     3.92      &     4.40      &     4.53      \\
		DINO ViT-B/16                            & No  &     15.66      &     16.97      &     16.83      &     4.57      &     5.21      &     5.37      \\ \midrule
		MAE ViT-B/16                             & Yes &      0.34      &      3.37      &      2.03      &     0.28      &     0.15      &     0.23      \\
		MAE ViT-B/16                             & No  &      2.34      &      4.02      &      3.73      &     0.62      &     0.77      &     0.69      \\
		MAE ViT-L/16                             & Yes &      0.35      &      0.08      &      0.28      &     0.01      &     0.01      &     0.02      \\
		MAE ViT-L/16                             & No  &      0.00      &      0.22      &      0.21      &     0.05      &     0.06      &     0.06      \\
		MAE ViT-H/14                             & Yes &      0.04      &      0.11      &      0.04      &     0.01      &     0.02      &     0.05      \\
		MAE ViT-H/14                             & No  &      0.06      &      0.10      &      0.16      &     0.06      &     0.06      &     0.06      \\ \bottomrule
	\end{tabular}
\end{table}

\section{Experiments}
\label{sec:p868624df5b1379ea230cf3b21a4260f2}

We evaluate our baseline on the proposed benchmark to examine how single-object category supervision supports multi-object scene understanding. We analyze the effects of visual representations, the number of training examples, and training-image backgrounds. For object detection, we also compare pipelines based on Adaptive MaskCut (AMC) and selective search proposals (SSP)~\cite{selectivesearch}. The main paper reports primary metrics, while additional metrics and implementation details are provided in the supplementary material.

\begin{figure*}[!t]
	\centering
	\begin{minipage}[t]{0.49\textwidth}
		\centering
		\includegraphics[width=\linewidth]
		{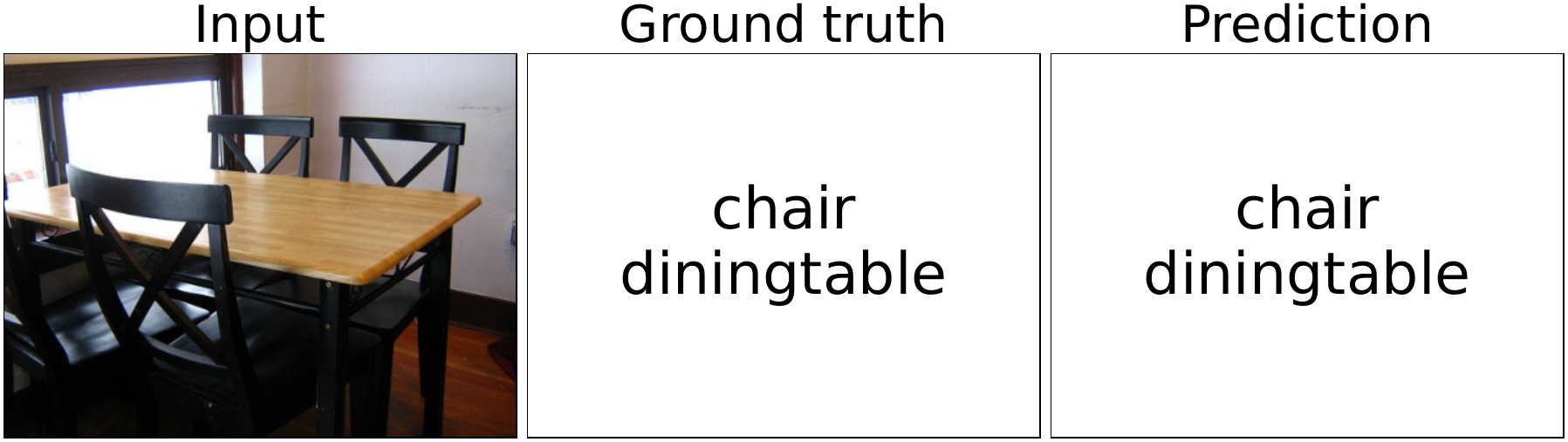}
		\par\small (a) Multi-label image classification
	\end{minipage}
	\hfill
	\begin{minipage}[t]{0.49\textwidth}
		\centering
		\includegraphics[width=\linewidth]
		{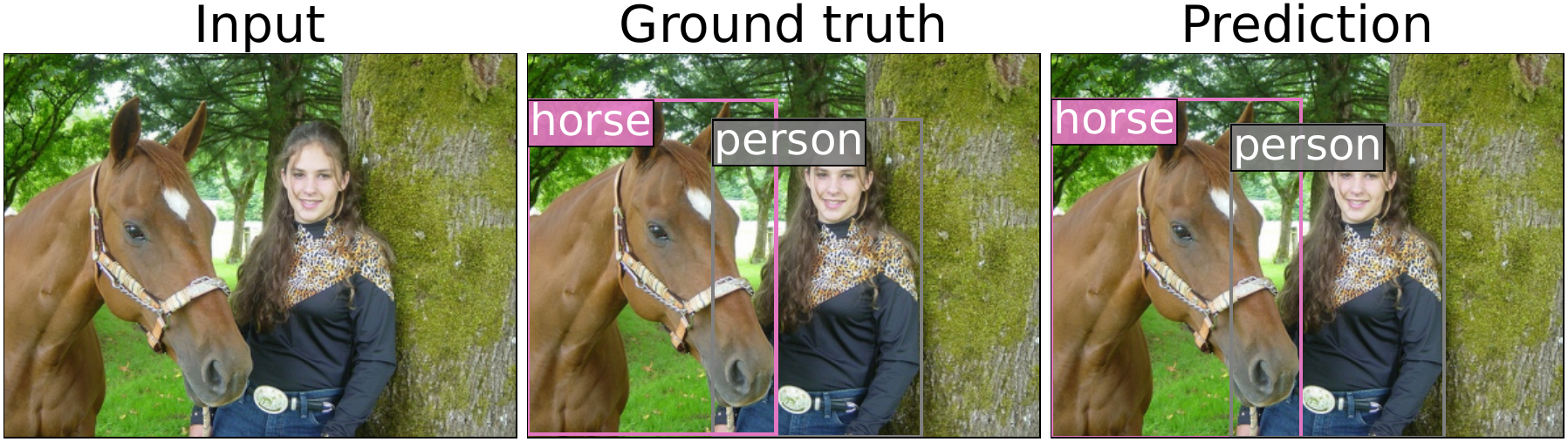}
		\par\small (b) Object detection
	\end{minipage}
	
	\vspace{6pt}
	
	\begin{minipage}[t]{0.49\textwidth}
		\centering
		\includegraphics[width=\linewidth]
		{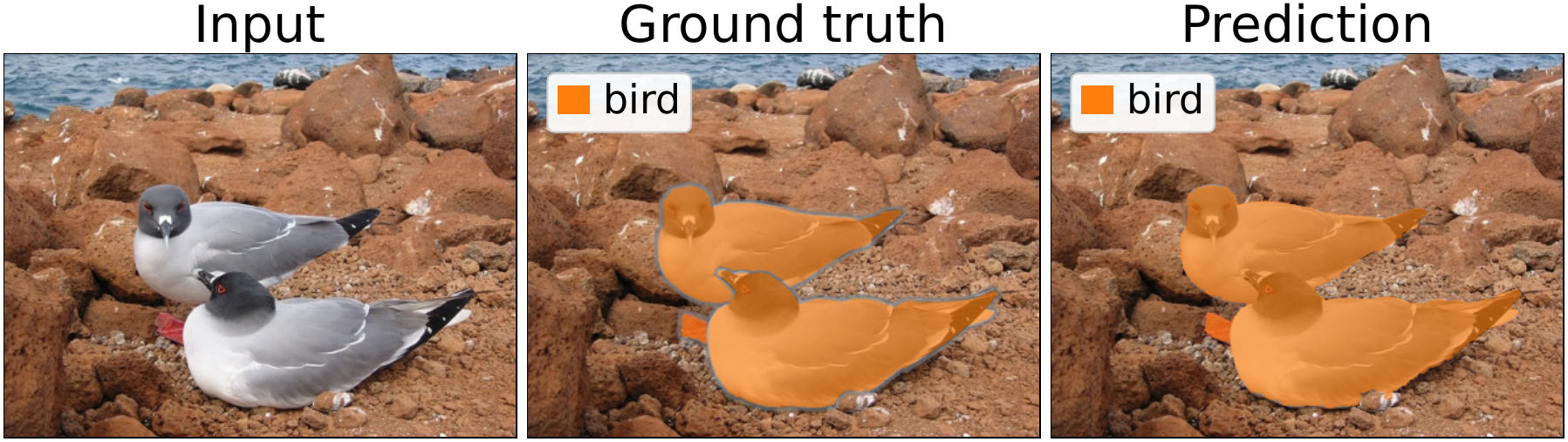}
		\par\small (c) Semantic segmentation
	\end{minipage}
	\hfill
	\begin{minipage}[t]{0.49\textwidth}
		\centering
		\includegraphics[width=\linewidth]
		{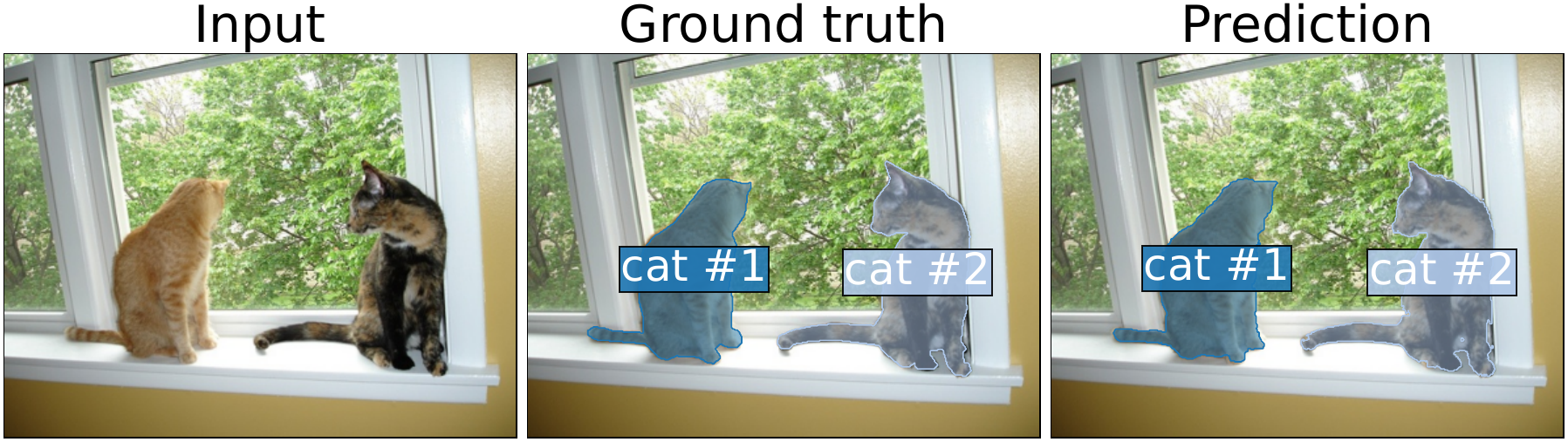}
		\par\small (d) Instance segmentation
	\end{minipage}
	
	\caption{
		Selected qualitative results on VOC using
		DINO ViT-S/8 with ten plain-background training
		images per category.
		Each panel shows the input, ground truth,
		and prediction from left to right.
	}
	\label{fig:qualitative_results}
\end{figure*}

\subsection{Setup}
\label{sec:experiment_setup}

\noindent\textbf{Implementation Details.}
We use publicly available self-supervised checkpoints without additional base set pretraining. By leveraging these community resources, our baseline requires no additional model training, facilitating reproduction and adoption. The encoders include DINO ViT-S/8, ViT-S/16, ViT-B/8, and ViT-B/16~\cite{dinov1}, and MAE ViT-B/16, ViT-L/16, and ViT-H/14~\cite{mae}. We adopt a few-shot experimental setting by constructing nested subsets of the benchmark's training set with 1, 5, or 10 single-object images per category to examine how the amount of category supervision affects performance on multi-object tasks.

\noindent\textbf{Image-Level Recognition.}
For MLIC, prototype similarities are converted into normalized category scores using a temperature-scaled softmax, followed by top-$p$ selection. We fix the temperature at $0.03$ for VOC\textsubscript{mlic}, COCO\textsubscript{mlic}, and VG500\textsubscript{mlic}, and at $0.1$ for NUS-WIDE\textsubscript{mlic}. For each combination of dataset, encoder, training-image background condition, and training set size, we select $p \in \{0.05,0.10,\ldots,0.90\}$ using validation label-set IoU and keep it fixed for testing.

\noindent\textbf{Region Generation.}
Our baseline uses AMC to generate candidate regions from local key features of the frozen encoder. The input side length is set to 480 and rounded upward to a multiple of the patch size. We set the patch-affinity similarity threshold to $\tau=0.2$, the weak affinity value to $\epsilon=10^{-5}$, and the minimum region-area ratio relative to the patch grid to $\eta=0.01$. Further details are provided in Section~\ref{sec:supp_stagehub} of the supplementary material. DenseCRF runs for 10 iterations, and we retain non-empty refined masks whose IoU with their coarse counterparts is at least $0.5$. Each retained mask defines a tight bounding box for cropping, and pixels outside the mask are set to black before region recognition.

\noindent\textbf{Task Readout.}
Each candidate region is assigned the category with the highest cosine similarity to the class prototypes, and this similarity serves as its prediction score. For OD, we retain predictions with scores of at least $0.2$. For SS, we apply the same threshold, assign overlapping pixels to the highest-scoring region, and label uncovered pixels as background. For IS, we retain all candidate instances without score thresholding. These post-processing rules remain fixed across training set sizes.

\subsection{Results across Tasks}
\label{sec:results}

Tables~\ref{tab:mlic_all}--\ref{tab:is_all} report results on the four tasks under one2multi learning. These results provide an initial reference for research on one2multi learning. Performance varies substantially across datasets and tasks, leaving considerable room for stronger methods. For object detection, we additionally compare the AMC-based baseline with SSP using the same encoder and class prototypes. AMC performs better across all evaluated DINO configurations, whereas SSP performs better with MAE ViT-L/16 and ViT-H/14. Because these pipelines also differ in crop masking and post-processing, this comparison evaluates their overall detection performance rather than proposal quality alone. Section~\ref{sec:discussion} further analyzes the effects of sample count, training backgrounds, and encoder choice.

\subsection{Qualitative Results}
Figure~\ref{fig:qualitative_results} illustrates how category knowledge acquired from single-object images supports different predictions in multi-object scenes. The baseline identifies co-occurring categories, localizes objects of different categories, and produces semantic and instance masks, including separate masks for objects of the same category. These selected examples illustrate the potential of one2multi learning without box or mask supervision on the single-object training images. Additional examples and selection criteria are provided in the supplementary material.

\subsection{Discussion}
\label{sec:discussion}

\begin{figure}[!t]
	\centering
	\includegraphics[width=0.7\linewidth]
	{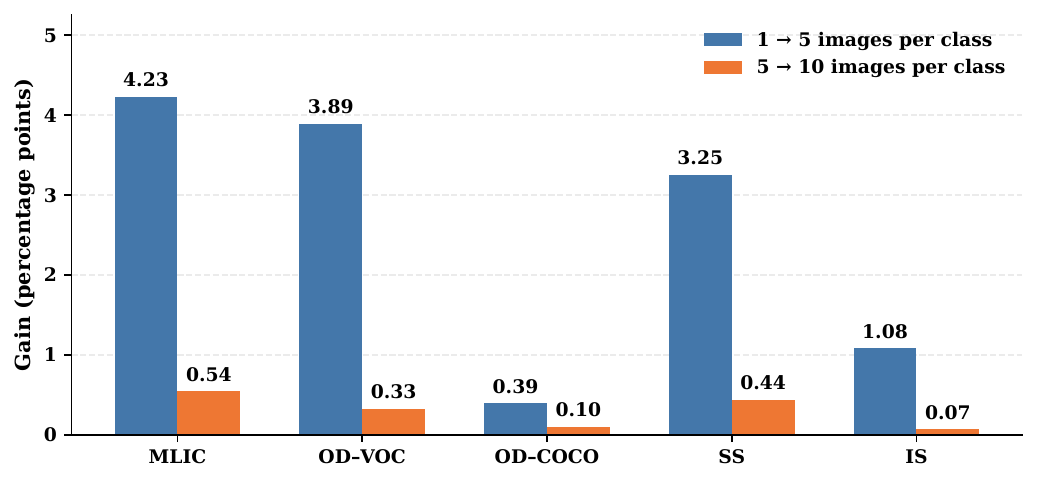}
	\caption{
		Performance gains from additional single-object
		training images.
	}
	\label{fig:sample_gains}
\end{figure}

\noindent\textbf{Diminishing Gains from Additional Training Examples.} Figure~\ref{fig:sample_gains} shows that increasing single-object training examples from one to five per category consistently improves average performance, whereas increasing from five to ten yields smaller gains. Multiple examples may produce more representative category prototypes by reducing sensitivity to individual appearances. However, additional category examples update only the prototypes in our baseline; they do not change the class-agnostic candidate regions and therefore cannot directly correct localization errors. These results echo the human-inspired motivation of one2multi learning, as a few labeled examples already yield substantial gains, although the diminishing returns observed here remain specific to our baseline.

\begin{figure}[!h]
	\centering
	\includegraphics[width=0.7\linewidth]
	{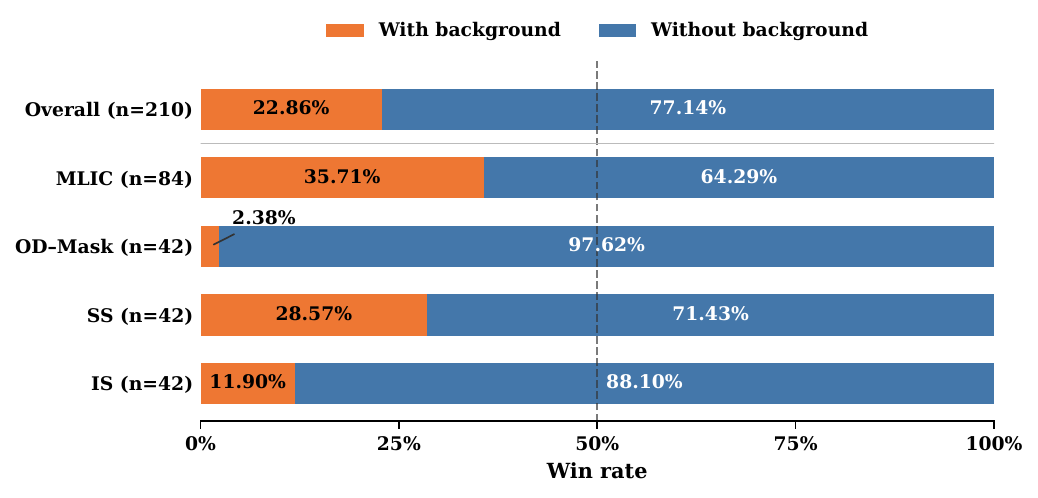}
	\caption{
		Paired win rates of training images with and without backgrounds. Each pair uses the same task, dataset, encoder, and training set size.
	}
	\label{fig:background_effect}
\end{figure}

\noindent\textbf{Plain Backgrounds Generally Yield Better Performance.}
As shown in Figure~\ref{fig:background_effect}, plain-background training images outperform contextual-background images in 162 of 210 matched configurations (77.14\%), with a particularly consistent advantage for AMC-based detection and instance segmentation. A single-object image does not necessarily yield an object-only representation: contextual features may also contribute to its category prototype. Plain backgrounds may reduce this influence, which could benefit recognition of region crops whose surrounding content is masked out. The weaker preference in MLIC suggests that the effect depends on how evaluation images are represented. However, the two training sets are generated independently, so differences in object appearance may also contribute. These results indicate a practical preference under our setup, rather than an isolated causal effect of background.

\begin{figure}[!htbp]
	\centering
	\includegraphics[width=0.7\linewidth]
	{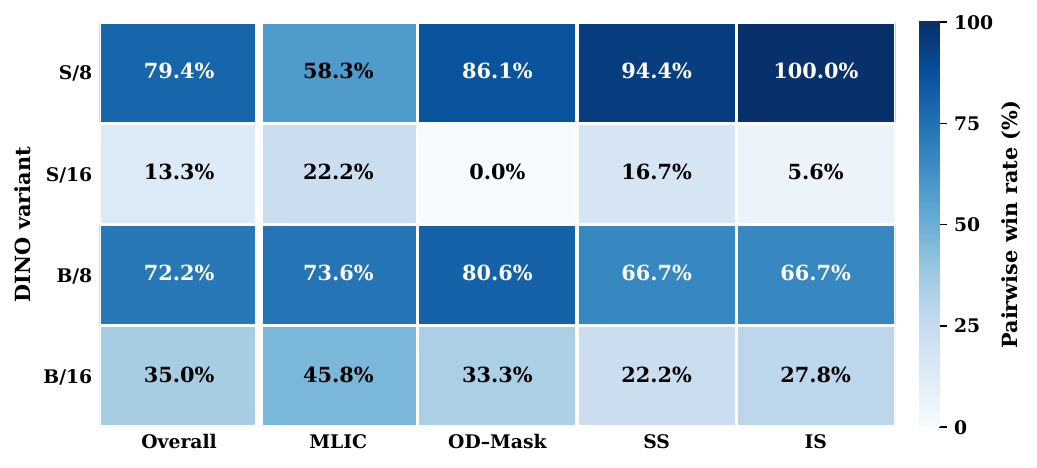}
	\caption{
		Pairwise win rates of DINO variants under
		matched task, dataset, background condition,
		and training set size.
	}
	\label{fig:encoder_comparison}
\end{figure}

\noindent\textbf{The Preferred Encoder Depends on the Task.}
Figure~\ref{fig:encoder_comparison} shows that DINO ViT-S/8 achieves the highest overall pairwise win rate and is particularly competitive on dense prediction tasks, whereas ViT-B/8 leads on MLIC. This difference suggests that features suited to global category matching may not be equally effective for class-agnostic localization. Smaller patches may benefit localization through finer spatial sampling, although our comparisons do not isolate this effect from changes in recognition features. These findings suggest that encoder selection should consider both category recognition and class-agnostic localization, rather than model size alone.

Overall, these observations suggest that effective one2multi learning depends not only on the number of labeled examples, but also on their visual content and the suitability of the representation for the target task. Additional examples and simpler backgrounds can improve category representations, while multi-object prediction also relies on effective localization. Our modular baseline provides a starting point for studying these complementary requirements.

\section{Conclusion}
\label{sec:conclusion}

We introduced one2multi learning, which investigates whether models can learn from a small number of single-object images with only category labels to perform multiple tasks in multi-object scenes. This setting offers a direction for reducing dependence on task-specific, densely annotated training data. We established a benchmark covering four fundamental vision tasks. We also proposed a unified baseline that decouples class-agnostic localization from category recognition. This modular design accommodates new categories by updating category knowledge while keeping the localization component unchanged. Experiments demonstrate its potential while highlighting substantial room for improvement. Future work includes extending the benchmark to more tasks and video understanding, and developing trainable methods that better exploit single-object supervision. 

\section*{Acknowledgments}
This work was supported by the National Science and
Technology Major Project on ``Brain Science and
Brain-Inspired Research'' under Grant No.~2026ZD0221404.

\bibliographystyle{unsrt}  
\bibliography{references}  

\clearpage

\appendix
\section{Benchmark Construction and Evaluation}
\label{sec:supp_one2multi}

This section supplements the benchmark described in Section~\ref{sec:paca1c0c0e6399d1a0ba69b42e744e62c}. We detail the data sources, training-image generation, task-specific annotation processing, and evaluation protocols. The four data components serve distinct purposes: the base set supports representation learning, the training sets provide category supervision, and the validation and test sets support hyperparameter selection and final evaluation, respectively.

\subsection{Datasets}
\label{sec:supp_o2m_datasets}

\paragraph{Base Set.}
We use ImageNet-1K~\cite{imagenet} as the base dataset and discard its category labels for representation learning. The original category annotations are not used for learning. As discussed in the main paper, ImageNet's object-centric imagery approximates the intended single-object learning material but does not strictly guarantee one object per image. The base set provides visual experience without establishing the target-category associations supplied by the training sets.

\paragraph{Training Sets.}
We construct synthetic training images for the category vocabularies of VOC~\cite{voc}, COCO~\cite{coco}, NUS-WIDE~\cite{nus}, and VG500~\cite{vg500}, comprising 20, 80, 81, and 500 categories, respectively. For each category, we generate 10 images with contextual backgrounds and 10 with plain backgrounds, yielding 20 images per category. The four training sets therefore contain 400, 1,600, 1,620, and 10,000 images, respectively, for a total of 13,620 images. Images are organized by dataset, background condition, and category. Only category labels are provided for learning, without bounding boxes, semantic masks, or instance masks.

\begin{figure*}[t]
	\centering
	\setlength{\tabcolsep}{3pt}
	\begin{tabular}{@{}cccc@{}}
		VOC: \textit{bicycle} &
		COCO: \textit{dog} &
		NUS-WIDE: \textit{fox} &
		VG500: \textit{bird} \\[3pt]
		
		\includegraphics[width=0.24\textwidth]
		{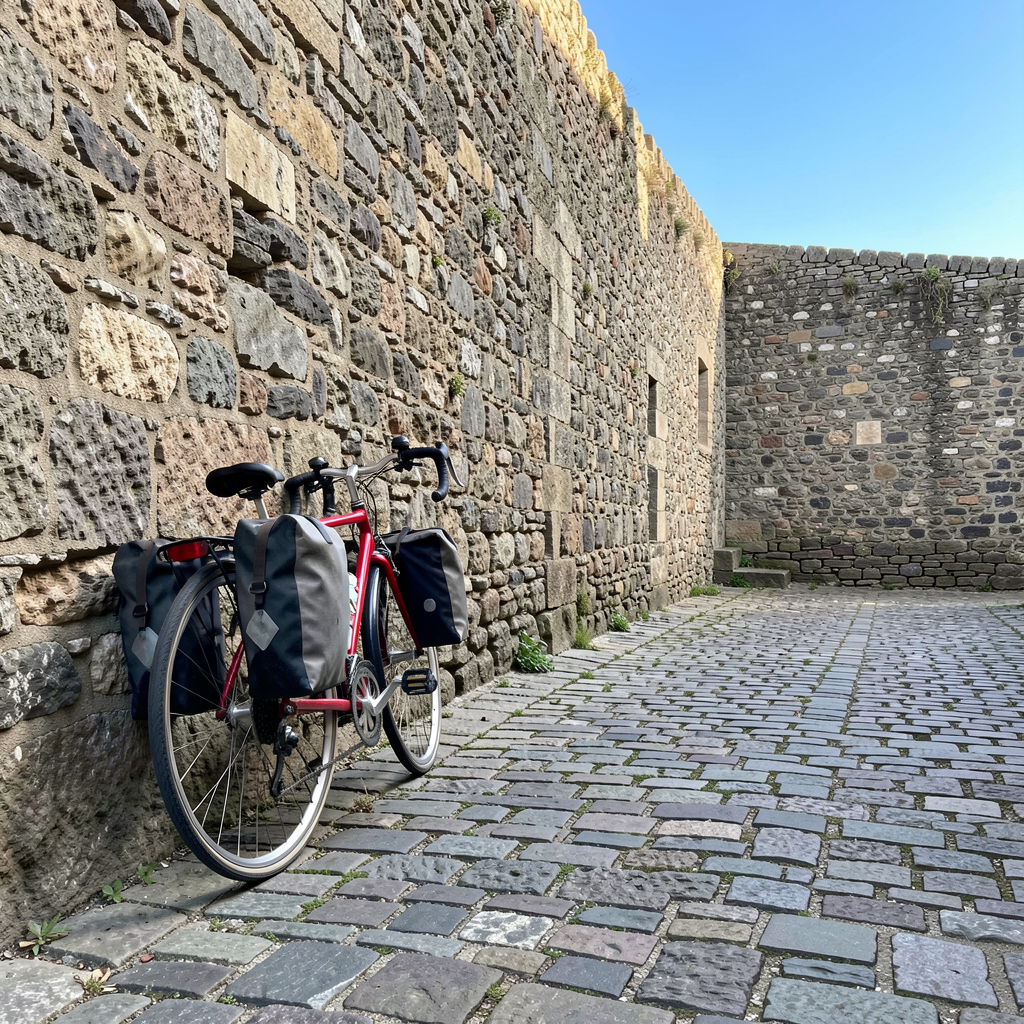} &
		\includegraphics[width=0.24\textwidth]
		{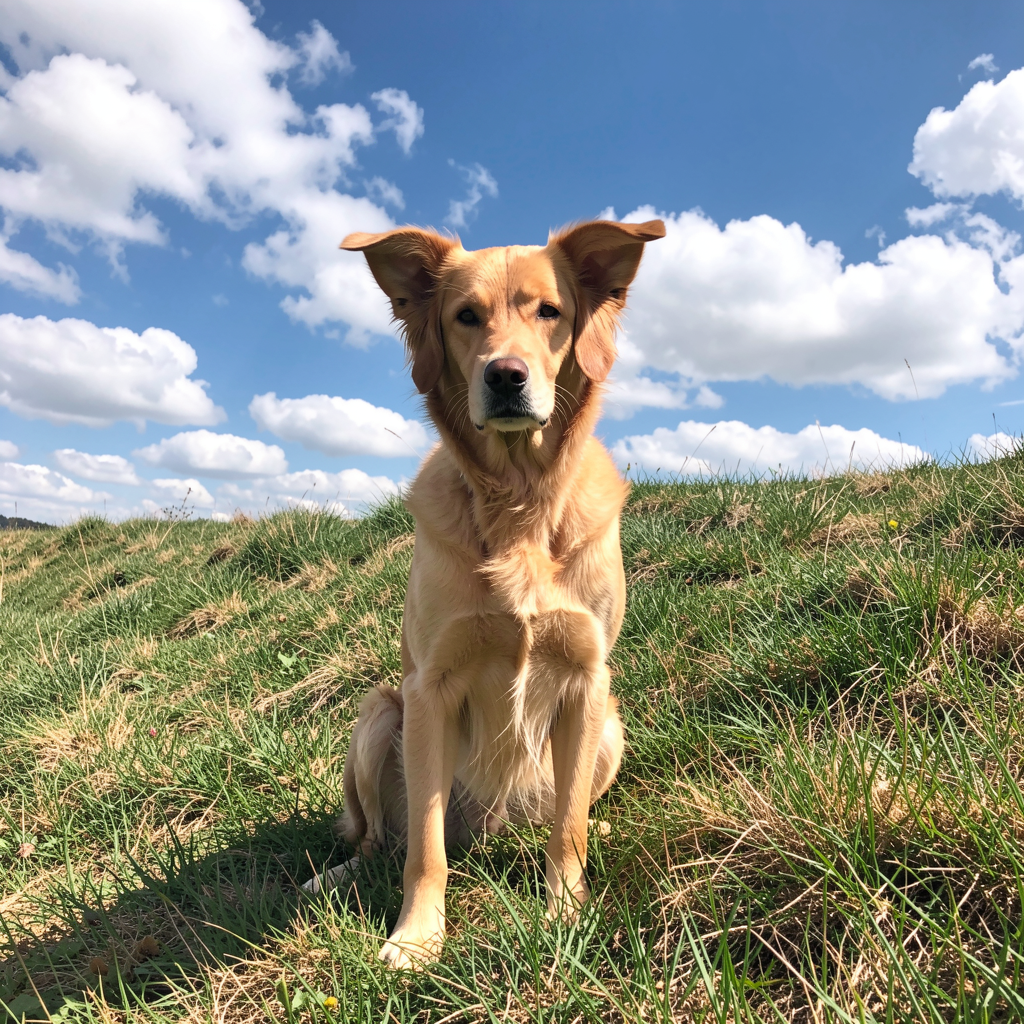} &
		\includegraphics[width=0.24\textwidth]
		{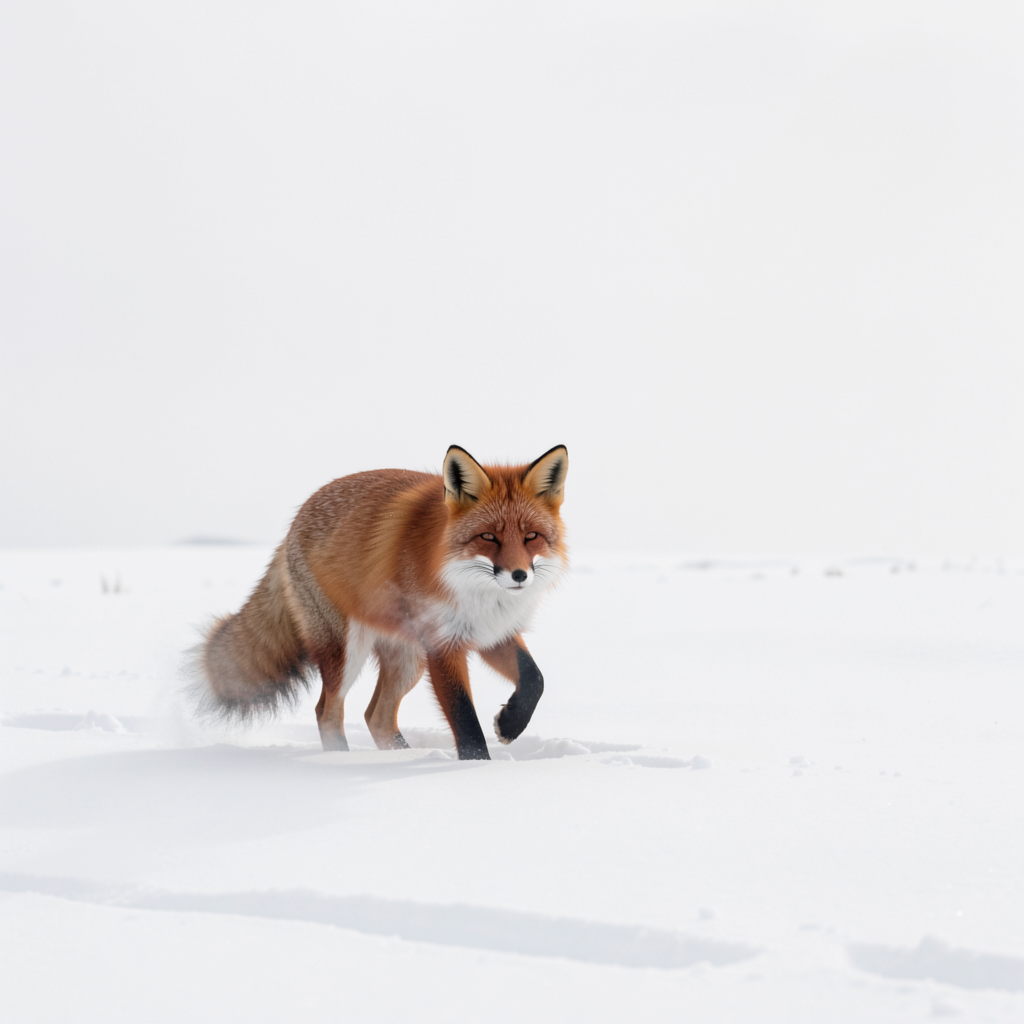} &
		\includegraphics[width=0.24\textwidth]
		{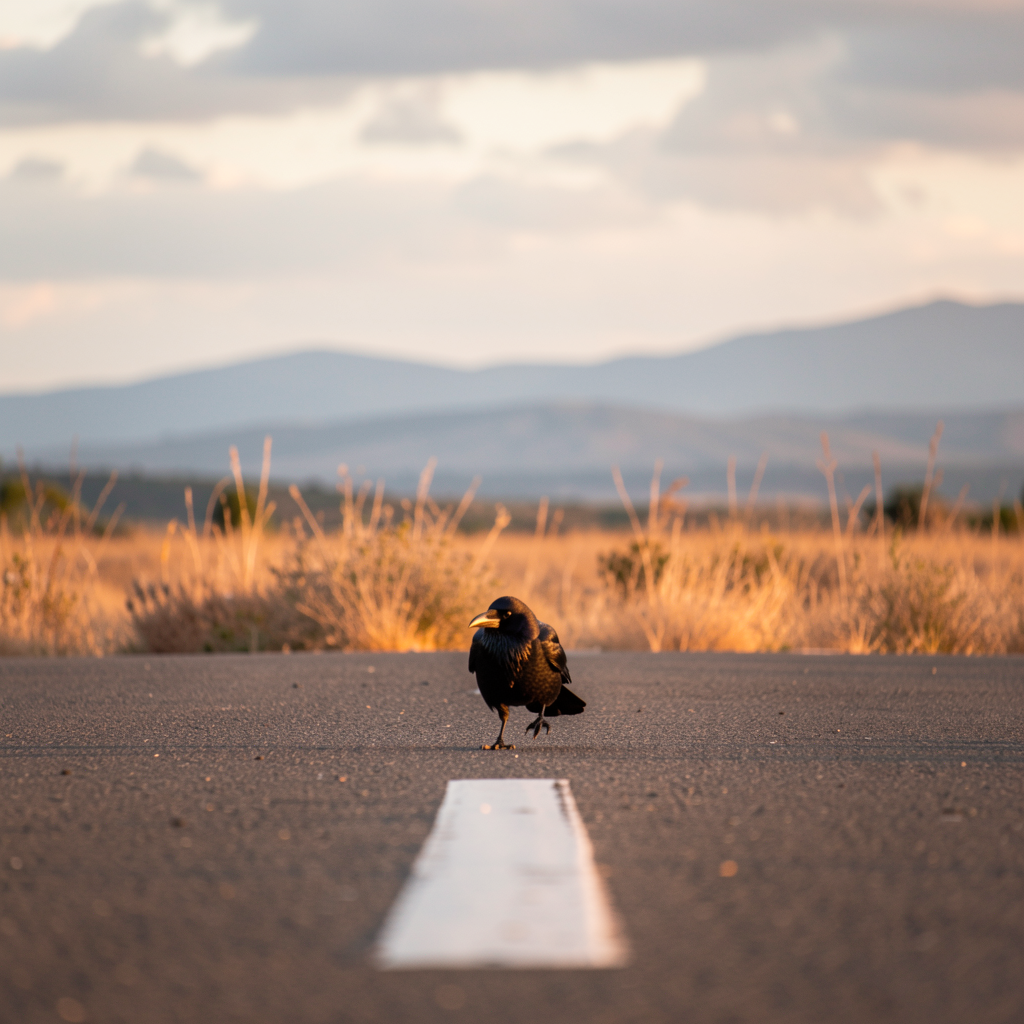} \\
		
		\multicolumn{4}{c}{(a) Contextual backgrounds} \\[6pt]
		
		\includegraphics[width=0.24\textwidth]
		{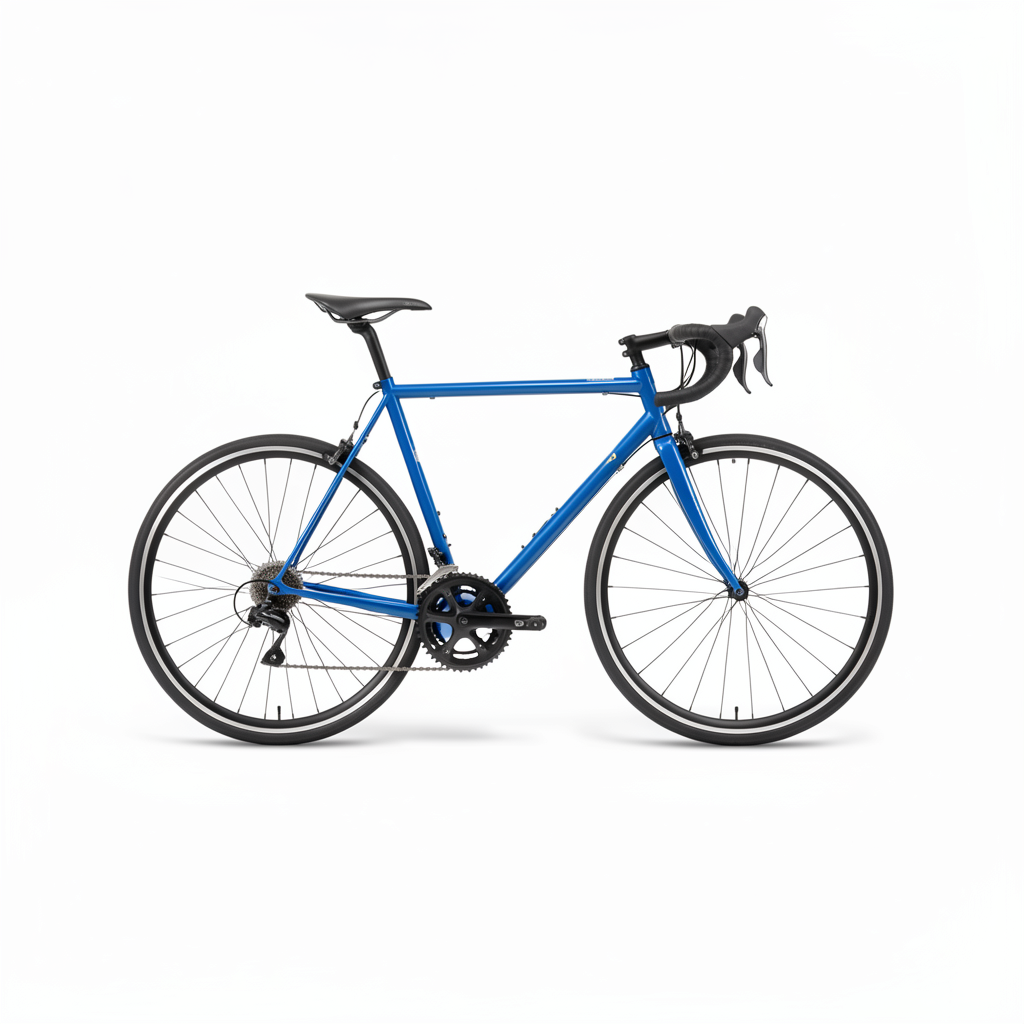} &
		\includegraphics[width=0.24\textwidth]
		{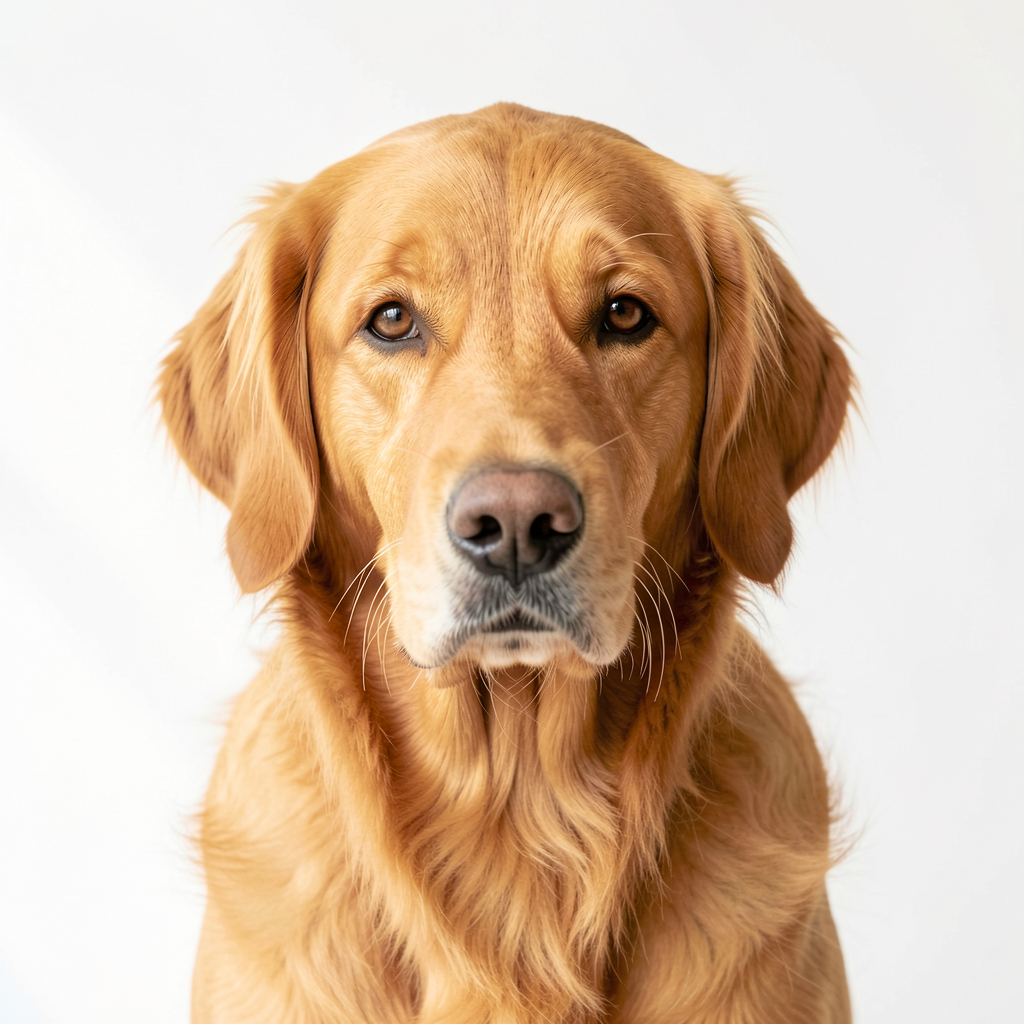} &
		\includegraphics[width=0.24\textwidth]
		{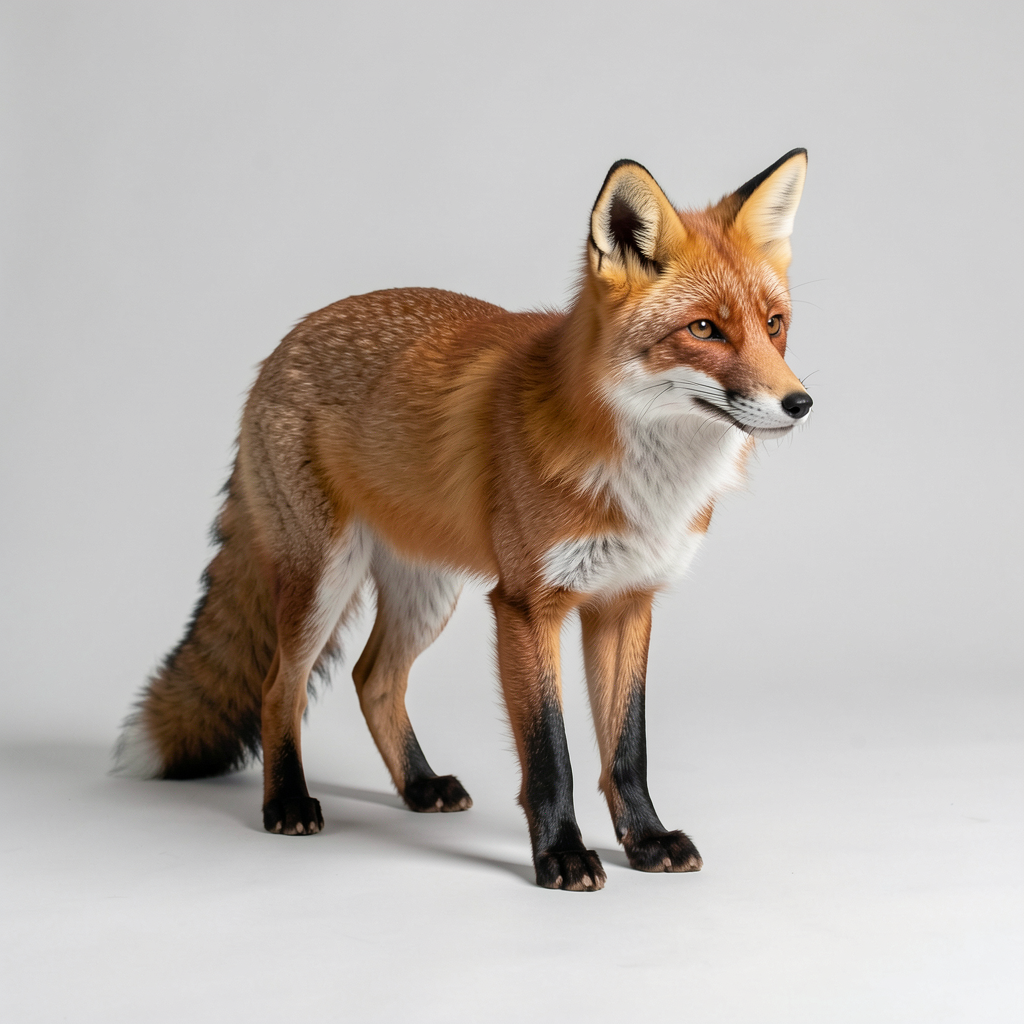} &
		\includegraphics[width=0.24\textwidth]
		{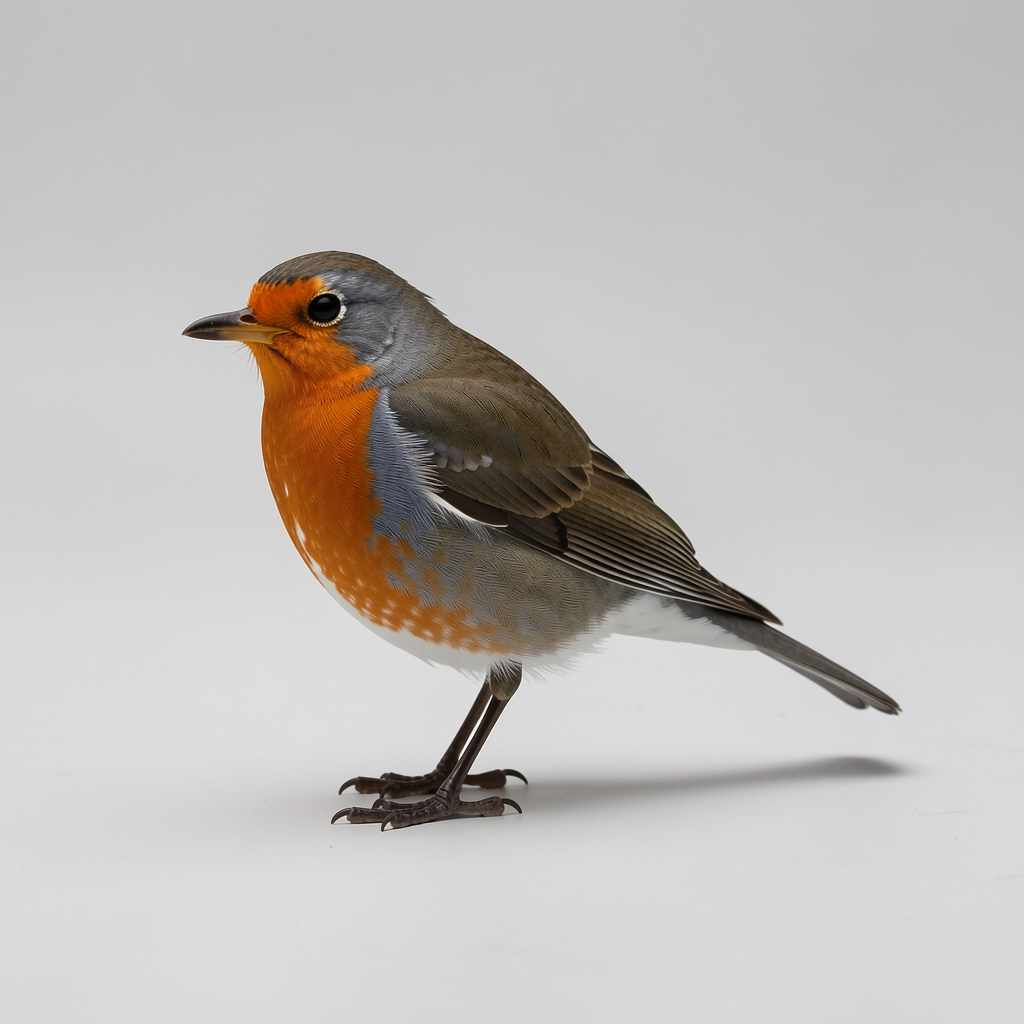} \\
		
		\multicolumn{4}{c}{(b) Plain backgrounds}
	\end{tabular}
	\caption{
		Synthetic training examples for four dataset vocabularies
		under contextual (top) and plain (bottom) backgrounds.
		Images in each column share the target category
		but are generated independently.
		Prompt-based constraints do not guarantee
		strict single-object content.
	}
	\label{fig:generated_training_examples}
\end{figure*}

\paragraph{Multi-Label Image Classification.} 
We evaluate MLIC on VOC\textsubscript{mlic}, COCO\textsubscript{mlic}, NUS-WIDE\textsubscript{mlic}, and VG500\textsubscript{mlic}, with validation and test splits summarized in Table~\ref{tab:supp_mlic_splits}. Each image is associated with a set of category labels from the corresponding dataset vocabulary. The task is to identify the categories present in an image, without predicting their locations or instance counts. The category vocabulary is shared with the corresponding synthetic single-object training set.

\begin{table}[t]
	\centering
	\caption{Validation and test splits for MLIC. Split sizes are reported as image counts.}
	\label{tab:supp_mlic_splits}
	\small
	\begin{tabular}{@{}lrrr@{}}
		\toprule
		Dataset                      & Classes & Validation &   Test \\ \midrule
		VOC\textsubscript{mlic}      &      20 &      2,510 &  4,952 \\
		COCO\textsubscript{mlic}     &      80 &     82,783 & 40,504 \\
		NUS-WIDE\textsubscript{mlic} &      81 &    119,103 & 50,720 \\
		VG500\textsubscript{mlic}    &     500 &     82,904 & 10,000 \\ \bottomrule
	\end{tabular}
\end{table}

\paragraph{Object Detection Datasets.}
Table~\ref{tab:supp_od_splits} summarizes the validation and test sets for OD. For VOC\textsubscript{od}, we use the official training and validation splits for validation and testing, respectively, rather than the official test split. For COCO\textsubscript{od}, we partition the original validation set into disjoint validation and test subsets. Both datasets provide category-labeled bounding boxes. VOC difficult-object flags are retained for evaluation.

\begin{table}[t]
	\centering
	\caption{Validation and test splits for OD. Split sizes are reported as image counts.}
	\label{tab:supp_od_splits}
	\small
	\begin{tabular}{@{}lrrr@{}}
		\toprule
		Dataset                & Classes & Validation &   Test \\ \midrule
		VOC\textsubscript{od}  &      20 &      2,501 &  2,510 \\
		COCO\textsubscript{od} &      80 &     20,000 & 20,504 \\ \bottomrule
	\end{tabular}
\end{table}

\paragraph{Semantic Segmentation.}
Table~\ref{tab:supp_seg_splits} summarizes the dataset splits shared by SS and IS. For VOC\textsubscript{ss}, we use the official segmentation validation and test splits with their semantic annotations, preserving background labels and ignored regions. COCO\textsubscript{ss} uses the same disjoint validation and test subsets as OD. Semantic targets are derived from instance annotations: masks of the same category are merged, while overlaps between different categories and crowd regions are ignored. Pixels outside annotated foreground masks are assigned background. This defines an instance-derived semantic segmentation protocol, rather than an official COCO semantic or panoptic evaluation.

\paragraph{Instance Segmentation.}
For VOC\textsubscript{is}, instance identities are obtained from the instance annotations, and each instance's category is determined from its corresponding semantic labels. Background and ignored regions are not treated as instances. For COCO\textsubscript{is}, we use the annotated instance masks, preserving category labels and crowd flags. Instances with empty rasterized masks are excluded.

\begin{table}[t]
	\centering
	\caption{Validation and test splits shared by SS and IS.
		Class counts exclude background.
		Split sizes are reported as image counts.}
	\label{tab:supp_seg_splits}
	\small
	\begin{tabular}{@{}lrrr@{}}
		\toprule
		Dataset & Classes & Validation & Test \\
		\midrule
		VOC & 20 & 213    & 210 \\
		COCO    & 80 & 20,000 & 20,504 \\
		\bottomrule
	\end{tabular}
\end{table}

\subsection{Training Image Generation}
\label{sec:supp_image_generation}

We follow the two-stage generation pipeline shown in Figure~\ref{fig:41182dc8}. A language model (deepseek-v4-flash~\cite{deepseekv41flash}) converts the target category, requested sample count, and background condition into descriptive prompts. A text-to-image model (FLUX.2-klein-9B~\cite{flux2klein9b}) then generates one image for each prompt. Table~\ref{tab:generation_settings} summarizes the configuration, and Table~\ref{tab:prompt_pipeline} provides concrete LLM inputs, generated prompts, and resulting images for the VOC category \textit{bicycle}. Figure~\ref{fig:generated_training_examples} presents generated examples from all four dataset vocabularies under both background conditions.

We generate contextual- and plain-background images separately. Contextual prompts specify realistic surroundings, while plain-background prompts specify neutral, studio-style, or minimally textured settings that may retain natural shadows and depth. Both conditions encourage variation in viewpoint, pose, scale, lighting, appearance, and composition. Each prompt requests one identifiable target instance and excludes recognizable objects from other categories in the corresponding dataset vocabulary. The target category serves as the image-level label, without bounding-box or mask annotations.

These constraints are imposed through prompting. The generation script verifies file integrity but does not check object count or category correctness, so successful generation does not guarantee compliance with the single-object requirement. The seed schedule and metadata fields in Table~\ref{tab:generation_settings} describe the current script, not a complete provenance record for every previously generated image.

\begin{table}[t]
	\centering
	\caption{Single-object training image generation settings.
		Object and background constraints are imposed through prompts.}
	\label{tab:generation_settings}
	\small
	\renewcommand{\arraystretch}{1.12}
	\begin{tabular}{@{}p{0.35\linewidth}p{0.59\linewidth}@{}}
		\hline
		Configuration & Setting \\
		\hline
		Prompt generator & deepseek-v4-flash \\
		Reasoning mode & Enabled; high reasoning effort \\
		Image generator & FLUX.2-klein-9B \\
		Resolution & $1024 \times 1024$ \\
		Inference steps & 4 \\
		Guidance scale & 1.0 \\
		Numerical precision & BF16 \\
		Random seeds & Sequential integers starting from 0
		within each generation run \\
		Images per prompt & 1 \\
		Background conditions & Contextual / plain \\
		Object constraint & Exactly one target instance;
		no recognizable objects from
		other dataset categories \\
		Visual style & Photorealistic \\
		Diversity factors & Viewpoint, pose, scale, lighting,
		appearance, and composition \\
		Supervision & Category labels only \\
		Output format & PNG with prompt, model, and seed metadata \\
		\hline
	\end{tabular}
\end{table}

\begin{table*}[t]
	\centering
	\caption{Two-stage prompting examples for the VOC category
		\textit{bicycle} under contextual and plain backgrounds.
		LLM inputs are excerpts with configuration values substituted.
		Image-generation prompts are reproduced verbatim,
		with the resulting images shown below.}
	\label{tab:prompt_pipeline}
	\footnotesize
	\renewcommand{\arraystretch}{1.1}
	\begin{tabular}{@{}p{0.18\textwidth}p{0.38\textwidth}p{0.38\textwidth}@{}}
		\hline
		Stage & Contextual background & Plain background \\
		\hline
		
		LLM system input &
		\multicolumn{2}{p{\dimexpr0.76\textwidth+2\tabcolsep\relax}@{}}{
			You are an AI image prompt generator specialized in
			creating high-quality prompts for synthetic image
			dataset generation.} \\
		\hline
		
		LLM user input:
		generation request &
		\multicolumn{2}{p{\dimexpr0.76\textwidth+2\tabcolsep\relax}@{}}{
			Please generate 10 high-quality English prompts for
			the single target label in the voc dataset.
			These prompts will be used with the
			black-forest-labs/FLUX.2-klein-9B text-to-image model
			to generate synthetic images for training an image
			classification model.
			\newline
			Target label: bicycle} \\
		\hline
		
		LLM user input:
		background condition &
		Background setting: with background.
		\newline
		Every prompt must place the target object in a
		realistic and contextually appropriate environment.
		The background should provide meaningful real-world
		variation but must not contain any recognizable object
		belonging to any other label in the dataset.
		&
		Background setting: without background.
		\newline
		Every prompt must place the target object against a
		clean, plain, neutral, studio-style, or minimally
		textured background. The image should remain
		photorealistic with natural lighting, shadows, and
		depth rather than appearing as an artificial cutout. \\
		\hline
		
		LLM user input:
		object constraint &
		\multicolumn{2}{p{\dimexpr0.76\textwidth+2\tabcolsep\relax}@{}}{
			Each image should contain exactly one clearly
			identifiable instance of the target category.
			If additional elements are unavoidable, they must
			be visually insignificant, unrecognizable, or
			reduced to non-object textures.} \\
		\hline
		
		LLM user input:
		output requirement &
		\multicolumn{2}{p{\dimexpr0.76\textwidth+2\tabcolsep\relax}@{}}{
			The prompts array must contain exactly 10 strings.
			Return valid JSON only. Do not wrap the JSON in
			Markdown code fences. Do not include numbering,
			comments, notes, or explanations outside the
			JSON object.} \\
		\hline
		
		LLM output:
		image prompt example &
		A single black bicycle with a slightly bent rear
		wheel is standing on wet asphalt during light rain,
		with a blurred empty road and a dark crosswalk
		behind it.
		&
		A full side view of a blue road bicycle with slim
		tires and drop handlebars standing upright on a
		seamless white studio background, with a soft natural
		shadow beneath the wheels and crisp photorealistic
		lighting. \\
		\hline
		
		Image generation &
		\multicolumn{2}{p{\dimexpr0.76\textwidth+2\tabcolsep\relax}@{}}{
			Each generated prompt is passed to FLUX.2-klein-9B
			to produce one $1024 \times 1024$ image using
			4 inference steps and a guidance scale of 1.0.} \\
		\hline
		
		Generated image &
		{\centering
			\includegraphics[width=0.75\linewidth]{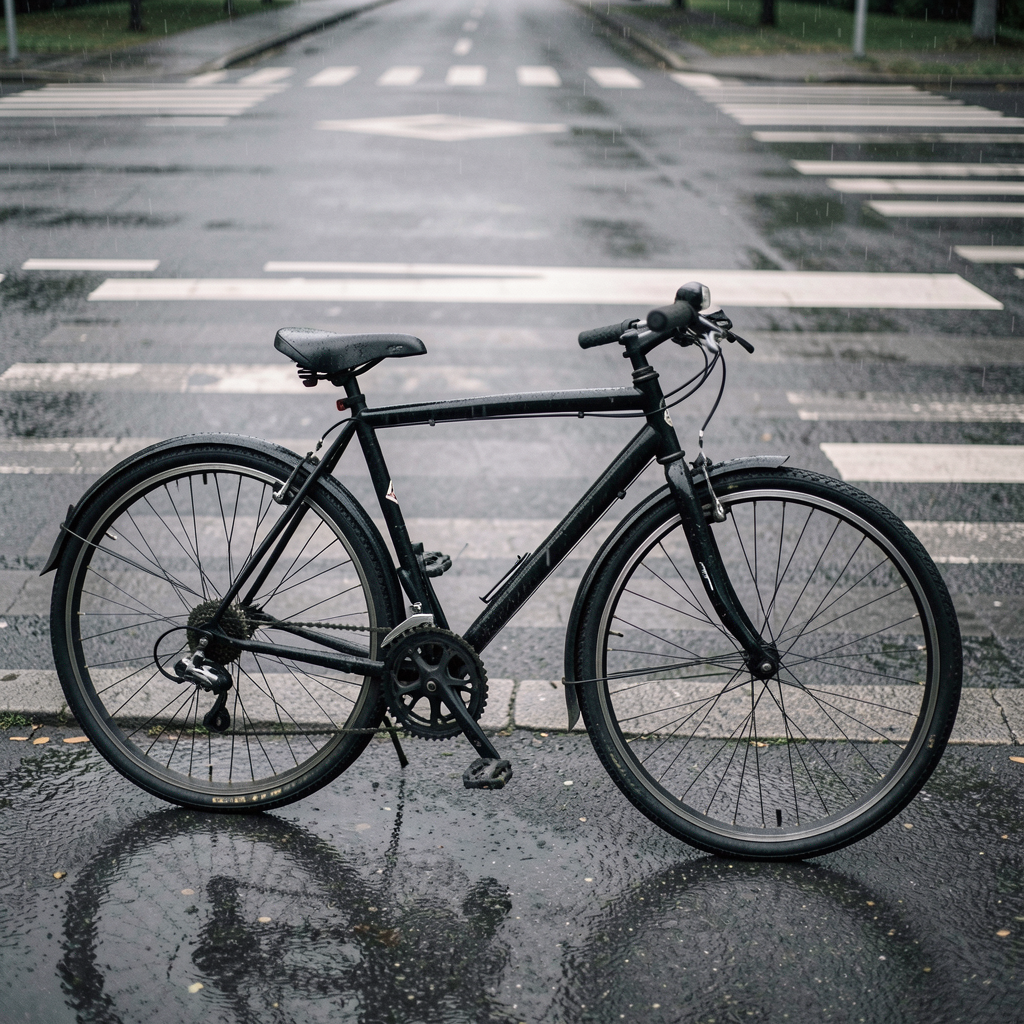}
			\par}
		&
		{\centering
			\includegraphics[width=0.75\linewidth]{figures/bicycle_plain.png}
			\par}
		\\
		\hline
	\end{tabular}
\end{table*}

\subsection{Evaluation Protocols}
\label{sec:supp_o2m_protocol}

\paragraph{Multi-Label Image Classification.}
Let $Y_i$ and $\widehat{Y}_i$ denote the ground-truth and predicted category sets for image $i$. Following the evaluation protocol in~\cite{mltr}, we use image-averaged label-set IoU as the primary metric:
\begin{equation}
	\mathrm{IoU}
	=
	\frac{1}{n_{\mathrm{eval}}}
	\sum_{i=1}^{n_{\mathrm{eval}}}
	\frac{|Y_i \cap \widehat{Y}_i|}
	{|Y_i \cup \widehat{Y}_i|},
\end{equation}
where $n_{\mathrm{eval}}$ is the number of evaluation images. We additionally compute image-averaged precision
and recall:
\begin{equation}
	\begin{aligned}
		\mathrm{Precision}
		&=
		\frac{1}{n_{\mathrm{eval}}}\sum_{i=1}^{n_{\mathrm{eval}}}
		\frac{|Y_i \cap \widehat{Y}_i|}
		{|\widehat{Y}_i|},\\
		\mathrm{Recall}
		&=
		\frac{1}{n_{\mathrm{eval}}}\sum_{i=1}^{n_{\mathrm{eval}}}
		\frac{|Y_i \cap \widehat{Y}_i|}
		{|Y_i|}.
	\end{aligned}
\end{equation}
When both sets are empty, all three image-level metrics are assigned a value of one. When only one set is empty, all three are assigned zero. These metrics average over images rather than categories.

\paragraph{Object Detection.}
For VOC\textsubscript{od}, we report class-averaged Box $\mathrm{AP}_{50}$ using all-point interpolation of the precision--recall curve. Detections matched to difficult ground-truth objects are ignored. For COCO\textsubscript{od}, we use the COCO bounding-box evaluator with its default IoU thresholds and maximum-detection settings. The primary metric is Box $\mathrm{AP}_{50:95}$, averaged over thresholds $\{0.50,0.55,\ldots,0.95\}$.

\paragraph{Semantic Segmentation.}
We accumulate a confusion matrix $\mathbf{H}$ over all valid pixels in the evaluation split. Rows index ground-truth classes and columns index predicted classes. For category $c$, 
\begin{equation} 
	\mathrm{IoU}_c = \frac{H_{cc}} {\sum_j H_{cj}+\sum_j H_{jc}-H_{cc}}. 
\end{equation} 
The reported mIoU averages over categories with nonzero union, including background. Ignored pixels are excluded before accumulating the confusion matrix. Thus, this metric is computed over the complete evaluation split rather than by averaging per-image mIoU.

We also report pixel accuracy (PA), the proportion of valid pixels assigned the correct class, and mean class accuracy (mAcc), the mean of $H_{cc}/\sum_j H_{cj}$ over classes with at least one ground-truth pixel. Both metrics are computed from the same accumulated confusion matrix and include background. Classes absent from the ground truth are excluded from mAcc.

\paragraph{Instance Segmentation.}
We use COCO-style Mask $\mathrm{AP}_{50:95}$ as the primary metric on both VOC\textsubscript{is} and COCO\textsubscript{is}. It averages mask AP over IoU thresholds from $0.50$ to $0.95$ in increments of $0.05$. For VOC\textsubscript{is}, pixels marked void in either the semantic or instance annotation are excluded from both predicted and ground-truth masks. Predictions lying entirely within these ignored regions are excluded, as are ground-truth instances with no remaining valid pixels. For COCO\textsubscript{is}, crowd flags are passed to the COCO-style evaluator for crowd-aware matching. Images without predictions remain part of the evaluation on both datasets.

\section{Baseline Details}
\label{sec:supp_baseline}
Our baseline separates class-agnostic localization from category recognition. \textbf{The motivation is that locating an object does not necessarily require knowing its category}. As illustrated in Figure~\ref{fig:intuition}, an unfamiliar object can be localized before category knowledge is acquired. We therefore use class-agnostic localization to identify candidate regions and category prototypes constructed from labeled single-object images to recognize them. This separation aligns with the human-inspired motivation of one2multi learning, where new category knowledge supports recognition in complex scenes while existing localization capabilities are reused. The modular design allows new categories to be incorporated by constructing additional prototypes, without retraining the visual encoder or localization component. Using publicly available self-supervised checkpoints also avoids additional model training in our implementation. The separation further allows localization and recognition components to be examined or replaced independently, providing a flexible reference for future one2multi methods.

This section provides implementation details for the baseline introduced in Section~\ref{sec:pcc914eda9010c895f25245151653b64c}. We describe self-supervised feature extraction, category prototypes, and StageHub.

\subsection{Self-Supervised Learning}
\label{sec:supp_encoders}

We use public self-supervised DINO~\cite{dinov1} and MAE~\cite{mae} checkpoints without additional base set pretraining. The encoder remains frozen and provides global CLS features for prototype construction and recognition, along with local patch features for class-agnostic localization. Images are converted to RGB and processed using the image processor associated with the selected checkpoint. Recognition uses the final-layer CLS feature. For MAE, the masking ratio is set to zero and deterministic ascending noise values preserve the original patch order.

\subsection{Prototypes}
\label{sec:supp_prototypes}

We follow Equation~\ref{eq:class_prototype} to construct the class prototypes. For category $c$, the CLS features of its training images $\mathfrak{D}_{o,c}$ are averaged to obtain $\boldsymbol{\mu}_c$, which is then normalized to produce $\mathbf{p}_c$. Individual training features are not normalized before averaging. A small positive lower bound on the norm is used in the implementation for numerical stability. The category-to-index mapping determines the ordering of the prototypes and is shared with the evaluation labels.

\subsection{StageHub}
\label{sec:supp_stagehub}

StageHub reuses the outputs of successive processing stages to support multiple tasks, as described in Section~\ref{sec}. The following sections specify these stages and their task readouts without introducing separately trained prediction heads.

\subsubsection{Image-Level Recognition}
\label{sec:supp_image_recognition}

Let the normalized global CLS feature be
\begin{equation}
	\mathbf{z}(\mathbf{x})
	=
	\frac{\mathbf{F}(\mathbf{x})[0,:]}
	{\|\mathbf{F}(\mathbf{x})[0,:]\|_2}.
\end{equation}
Its similarity to prototype $\mathbf{p}_c$ is
\begin{equation}
	s_c(\mathbf{x})
	=
	\mathbf{z}(\mathbf{x})^{\top}\mathbf{p}_c.
\end{equation}
A temperature-scaled softmax produces normalized category scores:
\begin{equation}
	\pi_c(\mathbf{x})
	=
	\frac{\exp(s_c(\mathbf{x})/T)}
	{\sum_{j=1}^{C}\exp(s_j(\mathbf{x})/T)},
\end{equation}
where $T>0$ and $C$ is the number of categories.

Let $\sigma$ order the categories by decreasing $\pi_c(\mathbf{x})$. Top-$p$ selection~\cite{top-p} retains
\begin{equation}
	L(\mathbf{x})
	=
	\min\left\{
	k\in\{1,\ldots,C\}:
	\sum_{j=1}^{k}\pi_{\sigma(j)}(\mathbf{x})
	\geq p
	\right\},
\end{equation}
yielding the label set $\widehat{Y}=\{\sigma(1),\ldots,\sigma(L(\mathbf{x}))\}$. This rule selects at least one category for $0<p\leq1$. The softmax scores are used for relative selection rather than as independent estimates of category presence.

\subsubsection{Adaptive MaskCut}
\label{sec:adamask}

\paragraph{Local Feature Extraction.}
We resize the input image to a square whose side length is rounded upward to a multiple of the encoder's patch size. After checkpoint-specific preprocessing, we extract key-projection features from the final attention layer, select the patch tokens excluding the CLS token, and normalize each patch feature. Let $\mathbf{k}_i$ denote the resulting feature at patch $i$, with $N$ patches in total.

\paragraph{Iterative Graph Partitioning.}
Let $\mathcal{C}_r$ denote the patches covered before iteration $r$. Previously covered features are suppressed: \begin{equation} 
	\widetilde{\mathbf{k}}_i^{(r)} = \begin{cases} \mathbf{0}, & i\in\mathcal{C}_r,\\ \mathbf{k}_i, & \text{otherwise}. \end{cases} 
\end{equation} All patches remain graph nodes. The affinity matrix is 
\begin{equation} 
	A_{ij}^{(r)} = \begin{cases} 1, & (\widetilde{\mathbf{k}}_i^{(r)})^{\top} \widetilde{\mathbf{k}}_j^{(r)}>\tau,\\ \epsilon, & \text{otherwise}, \end{cases} 
\end{equation} where $\tau$ is the similarity threshold and $\epsilon>0$ provides weak connections. With $D_{ii}^{(r)}=\sum_j A_{ij}^{(r)}$, normalized cuts~\cite{normalizedcuts} solve 
\begin{equation} 
	(\mathbf{D}^{(r)}-\mathbf{A}^{(r)}) \mathbf{y}^{(r)} = \lambda\mathbf{D}^{(r)}\mathbf{y}^{(r)}. 
\end{equation} 
The implementation solves the equivalent symmetric normalized Laplacian problem and converts the eigenvector back to this generalized form.

We threshold the second-smallest eigenvector at its mean. The partition is reversed if at least three grid corners are foreground, or if fewer than three corners are foreground but the largest absolute eigenvector response lies outside the foreground. After orientation correction, the four-connected foreground component containing the maximum eigenvector value defines the candidate region $\mathcal{M}_r$.

\paragraph{Adaptive Extraction and Stopping.}
We remove previously covered patches and retain the largest connected component:
\begin{equation}
	\mathcal{U}_r = \operatorname{LCC} (\mathcal{M}_r \setminus \mathcal{C}_r),
\end{equation}
where $\operatorname{LCC}$ retains the largest four-connected component on the patch grid. The minimum accepted region area is 
\begin{equation} 
	a_{\min}=\max(1,\lceil\eta N\rceil), 
\end{equation} 
where $N$ is the total number of patches and $\eta$ specifies the minimum area ratio relative to the complete patch grid. If $|\mathcal{U}_r|\geq a_{\min}$, the region is accepted and coverage is updated:
\begin{equation}
	\mathcal{C}_{r+1} = \mathcal{C}_r\cup\mathcal{U}_r.
\end{equation}

Extraction stops when the new region is smaller than $a_{\min}$, fewer than $\max(2,a_{\min})$ patches remain uncovered, or no valid partition is found. Uniform off-diagonal affinities and effectively constant eigenvectors are treated as invalid partitions. These conditions adapt the proposal count to the image rather than imposing a fixed number of masks.

\paragraph{Mask Refinement.}
Each accepted patch-level region is resized to the original image resolution using nearest-neighbor interpolation, producing a coarse mask $\overline{M}_r$. DenseCRF~\cite{densecrf} uses the original RGB image to produce a refined mask $M_r$. We retain nonempty masks satisfying
\begin{equation}
	\operatorname{IoU}(M_r,\overline{M}_r)
	=
	\frac{|M_r\cap\overline{M}_r|}
	{|M_r\cup\overline{M}_r|}
	\geq \rho.
\end{equation}
Coverage is updated before refinement. Consequently, a coarse region remains suppressed even if its refined mask is rejected.

The CRF unary probabilities assign foreground probability 0.9 inside the coarse mask and 0.1 outside it. The Gaussian pairwise term uses spatial scale 3 and compatibility 3. The bilateral term uses spatial scale 50, RGB scale 10, and compatibility 5. We use 10 CRF iterations and $\rho=0.5$. The nominal input side length is 480, rounded upward to a multiple of the patch size, with $\tau=0.2$, $\epsilon=10^{-5}$, and $\eta=0.01$, matching Section~\ref{sec:experiment_setup}.

\subsubsection{Bounding Box Generation}
\label{sec:bbg}

For a retained mask $M_r$, let $\Omega_r=\{(u,v):M_r(v,u)=1\}$. Its tight bounding box is defined by
\begin{equation}
	\begin{aligned}
		u_r^{-}&=\min_{(u,v)\in\Omega_r}u,
		&
		v_r^{-}&=\min_{(u,v)\in\Omega_r}v,
        &
		u_r^{+}&=1+\max_{(u,v)\in\Omega_r}u,
		&
		v_r^{+}&=1+\max_{(u,v)\in\Omega_r}v.
	\end{aligned}
\end{equation}
The right and bottom boundaries are exclusive.
Boxes are stored as
\begin{equation}
	b_r=
	(u_r^{-},v_r^{-},
	u_r^{+}-u_r^{-},v_r^{+}-v_r^{-}).
\end{equation}
These boxes serve both as localization outputs
and as crop boundaries for region recognition.

\subsubsection{Region Recognition}
\label{sec:supp_region_recognition}

For each mask $M_r$ and box $b_r$, we construct a region image
\begin{equation}
	\mathbf{x}_r
	=
	\operatorname{Crop}
	(\mathbf{x}\odot M_r,b_r),
\end{equation}
where the mask is broadcast across RGB channels. Pixels outside the mask are set to black before encoder preprocessing.

The frozen encoder extracts the crop's CLS feature, which is normalized to obtain $\mathbf{z}_r$. Prototype matching assigns
\begin{equation}
	\widehat{c}_r
	=
	\underset{c}{\arg\max}\;
	\mathbf{z}_r^{\top}\mathbf{p}_c,
	\qquad
	q_r
	=
	\max_c\mathbf{z}_r^{\top}\mathbf{p}_c.
\end{equation}
The category $\widehat{c}_r$ and score $q_r$ are attached to both $M_r$ and $b_r$. Here, $q_r$ is a similarity score rather than a calibrated probability.

\subsubsection{Task Readout}
\label{sec:supp_postprocessing}

The task specification determines the output format without modifying the encoder or prototypes. MLIC returns the selected image-level labels. OD returns classified boxes with scores $q_r\geq0.2$, whereas IS returns all retained instance masks with their categories and scores, without additional classification-score filtering. For SS, each pixel is assigned the class of the highest-scoring retained mask covering it:
\begin{equation}
	\widehat{y}_{\mathrm{SS}}(v,u)=
	\begin{cases}
		\widehat{c}_{r^\ast}, & \mathcal{R}(u,v)\neq\varnothing,\\
		0, & \text{otherwise},
	\end{cases}
\end{equation}
where $\mathcal{R}(u,v)=\{r:q_r\geq0.2,\ M_r(v,u)=1\}$ and $r^\ast=\arg\max_{r\in\mathcal{R}(u,v)}q_r$. Ties are resolved by proposal order, and 0 denotes background. All four tasks thus share the same intermediate representations without separately trained task-specific heads.

\section{Additional Experimental Results}
\label{sec:supp_experimental_results}

This section expands the cross-task results in Section~\ref{sec:results} with per-configuration tables and complementary metrics. The primary metrics assess performance under one2multi learning, and the additional metrics characterize label selection and segmentation behavior. Dataset splits and evaluation protocols follow Sections~\ref{sec:supp_o2m_datasets} and~\ref{sec:supp_o2m_protocol}. The encoder, prototype construction, and task readouts follow Section~\ref{sec:experiment_setup} of the main paper and Section~\ref{sec:supp_baseline} of this supplement.

\subsection{Multi-Label Image Classification}
\label{sec:supp_mlic_results}

Tables~\ref{tab:mlic_voc}--\ref{tab:mlic_vg500} report image-averaged label-set IoU, precision (P), and recall (R). The temperature is fixed at the dataset-specific value specified in the main paper. Only the top-$p$ threshold is selected by validation label-set IoU for each configuration. The reported P and R are evaluated at this same threshold, and they are not independently optimized on the validation or test set.

\paragraph{VOC\textsubscript{mlic}.} Table~\ref{tab:mlic_voc} demonstrates the potential of single-object category supervision for multi-label recognition. With plain-background training images, DINO ViT-B/16 improves from 56.24\% IoU at 1-shot to 65.70\% at 5-shot and 66.50\% at 10-shot. Most of the improvement occurs within the first five examples per category. At 10-shot, precision reaches 82.02\%, while recall remains at 67.56\%, indicating that the predicted labels are relatively reliable but do not fully cover the categories present. 

\paragraph{COCO\textsubscript{mlic}.} Table~\ref{tab:mlic_coco} shows that plain-background training images do not consistently provide better results. For ten-shot DINO ViT-B/8, plain backgrounds increase precision from 69.70\% to 70.74\%, but reduce recall from 45.02\% to 41.29\% and IoU from 38.64\% to 37.51\%. Higher precision therefore does not necessarily improve label-set overlap when category coverage decreases. These differences reflect both the training images and the top-$p$ threshold selected for each configuration. 

\paragraph{NUS-WIDE\textsubscript{mlic}.} Table~\ref{tab:mlic_nus} highlights the importance of interpreting recall alongside precision and IoU. One-shot MAE ViT-L/16 with contextual backgrounds achieves 74.39\% recall, but only 3.19\% precision and 3.15\% IoU, showing that high coverage can coexist with extensive false-positive predictions. The best ten-shot IoU is 20.28\%, obtained by DINO ViT-S/16 with plain backgrounds, although its recall is only 21.03\%. Accurate and comprehensive label prediction thus remains challenging under the current protocol. 

\paragraph{VG500\textsubscript{mlic}.} Table~\ref{tab:mlic_vg500} reveals substantial limitations of global prototype matching on the 500-category vocabulary. DINO ViT-B/8 with contextual backgrounds achieves the best ten-shot IoU of 7.11\%, compared with 7.05\% at 5-shot, while recall remains at 18.58\%. Increasing the number of category examples alone therefore provides limited improvement in this configuration. The results motivate stronger category representations and label-selection methods, although the difficulty cannot be attributed to vocabulary size alone.

\begin{table*}[t]
	\centering
	\caption{MLIC results (\%) on VOC.
		IoU, P, and R denote image-averaged label-set IoU,
		precision, and recall.
		Bold marks the highest value in each column.}
	\label{tab:mlic_voc}
	\small
	\setlength{\tabcolsep}{4pt}
	\begin{tabular}{llccccccccc}
		\toprule
		                                                               &     &            \multicolumn{3}{c}{1-shot}            &            \multicolumn{3}{c}{5-shot}            &           \multicolumn{3}{c}{10-shot}            \\ \cmidrule
		(lr){3-5}
		\cmidrule(lr){6-8}
		\cmidrule(lr){9-11}
		Encoder & BG  &      IoU       &       P        &       R        &      IoU       &       P        &       R        &      IoU       &       P        &       R        \\ \midrule
		DINO ViT-S/8                                                   & Yes &     51.64      &     64.37      &     51.82      &     63.28      &     78.08      &     64.13      &     63.54      &     78.18      &     66.22      \\
		DINO ViT-S/8                                                   & No  &     56.71      &     70.66      &     59.66      &     64.59      &     79.98      &     66.23      &     66.06      &     80.75      & \textbf{70.92} \\
		\addlinespace
		DINO ViT-S/16                                  & Yes &     47.27      &     57.90      &     52.68      &     61.20      &     74.43      &     65.89      &     61.42      &     74.56      &     66.25      \\
		DINO ViT-S/16                                                  & No  &     55.35      &     68.98      &     55.46      &     64.68      &     80.27      &     65.33      &     65.53      &     81.00      &     66.64      \\
		\addlinespace
		DINO ViT-B/8                                   & Yes &     49.58      &     62.45      &     49.79      &     63.38      &     78.88      &     64.19      &     63.87      &     79.04      &     66.83      \\
		DINO ViT-B/8                                                   & No  & \textbf{57.02} & \textbf{71.67} &     57.89      &     65.39      & \textbf{81.36} &     65.43      &     66.02      &     81.49      &     68.75      \\
		\addlinespace
		DINO ViT-B/16                                  & Yes &     49.18      &     61.18      &     50.40      &     61.18      &     75.40      &     63.60      &     62.16      &     76.78      &     64.47      \\
		DINO ViT-B/16                                                  & No  &     56.24      &     70.07      & \textbf{60.14} & \textbf{65.70} &     81.07      & \textbf{67.66} & \textbf{66.50} & \textbf{82.02} &     67.56      \\ \midrule
		MAE ViT-B/16                                                   & Yes &     14.08      &     17.61      &     17.26      &     25.86      &     32.21      &     25.86      &     25.74      &     31.74      &     25.74      \\
		MAE ViT-B/16                                                   & No  &     17.23      &     20.27      &     17.23      &     25.77      &     31.28      &     25.77      &     26.19      &     31.44      &     26.19      \\
		\addlinespace
		MAE ViT-L/16                                   & Yes &     12.84      &     15.71      &     21.48      &     18.81      &     23.50      &     22.02      &     21.14      &     25.94      &     26.52      \\
		MAE ViT-L/16                                                   & No  &     10.12      &     13.00      &     10.12      &     24.63      &     31.50      &     26.22      &     23.22      &     28.68      &     25.26      \\
		\addlinespace
		MAE ViT-H/14                                   & Yes &     10.72      &     14.64      &     12.66      &     13.23      &     16.67      &     17.38      &     14.44      &     18.13      &     15.90      \\
		MAE ViT-H/14                                                   & No  &      7.42      &      8.33      &     31.10      &     12.80      &     16.47      &     15.69      &     14.57      &     17.96      &     17.91      \\ \bottomrule
	\end{tabular}
\end{table*}

\begin{table*}[t]
	\centering
	\caption{MLIC results (\%) on COCO.
		IoU, P, and R denote image-averaged label-set IoU,
		precision, and recall.
		Bold marks the highest value in each column.}
	\label{tab:mlic_coco}
	\small
	\setlength{\tabcolsep}{4pt}
	\begin{tabular}{llccccccccc}
		\toprule
		& & \multicolumn{3}{c}{1-shot}
		& \multicolumn{3}{c}{5-shot}
		& \multicolumn{3}{c}{10-shot} \\
		\cmidrule(lr){3-5}
		\cmidrule(lr){6-8}
		\cmidrule(lr){9-11}
		Encoder & BG
		& IoU & P & R
		& IoU & P & R
		& IoU                         & P              & R              \\
		\midrule
		DINO ViT-S/8 & Yes
		& 32.73 & 58.07 & 39.49
		& \textbf{37.03} & 65.94 & 43.74
		& \textbf{38.70}              & 69.00          & \textbf{45.44} \\
		DINO ViT-S/8 & No
		& 33.48 & 60.78 & 39.90
		& 36.00 & 65.63 & 42.08
		& 36.93                       & 67.49          & 43.24          \\
		\addlinespace
		DINO ViT-S/16 & Yes
		& 28.56 & 49.11 & 36.72
		& 33.32 & 57.73 & 40.96
		& 35.69                       & 61.93          & 43.84          \\
		DINO ViT-S/16 & No
		& 31.14 & 58.06 & 36.03
		& 34.06 & 62.98 & 39.33
		& 35.34                       & 65.79          & 40.60          \\
		\addlinespace
		DINO ViT-B/8 & Yes
		& 32.79 & 57.90 & 40.72
		& 36.98 & 65.98 & \textbf{44.54}
		& 38.64                       & 69.70          & 45.02          \\
		DINO ViT-B/8 & No
		& \textbf{33.61} & \textbf{63.45} & 36.73
		& 36.04 & \textbf{66.89} & 40.96
		& 37.51                       & \textbf{70.74} & 41.29          \\
		\addlinespace
		DINO ViT-B/16 & Yes
		& 30.26 & 52.24 & 37.80
		& 35.06 & 60.76 & 42.82
		& 36.58                       & 63.83          & 44.50          \\
		DINO ViT-B/16 & No
		& 33.27 & 59.84 & \textbf{41.09}
		& 35.62 & 65.80 & 40.51
		& 36.96                       & 67.47          & 43.35          \\
		\midrule
		MAE ViT-B/16 & Yes
		& 6.39 & 10.57 & 10.25
		& 9.69 & 14.24 & 18.54
		& 11.81                       & 16.99          & 23.29          \\
		MAE ViT-B/16 & No
		& 7.78 & 11.74 & 13.40
		& 8.45 & 12.14 & 16.93
		& 10.02                       & 14.54          & 20.01          \\
		\addlinespace
		MAE ViT-L/16 & Yes
		& 5.34 & 9.67 & 7.29
		& 7.99 & 14.73 & 10.13
		& 9.93                        & 16.03          & 13.92          \\
		MAE ViT-L/16 & No
		& 5.68 & 9.78 & 8.24
		& 7.81 & 13.42 & 9.73
		& 8.15                        & 13.45          & 10.61          \\
		\addlinespace
		MAE ViT-H/14 & Yes
		& 4.06 & 5.80 & 10.72
		& 6.92 & 11.21 & 13.29
		& 7.87                        & 12.20          & 13.98          \\
		MAE ViT-H/14 & No
		& 4.31 & 4.87 & 37.97
		& 5.68 & 10.45 & 8.11
		& 5.81                        & 9.55           & 9.06           \\
		\bottomrule
	\end{tabular}
\end{table*}

\begin{table*}[t]
	\centering
	\caption{MLIC results (\%) on NUS-WIDE.
		IoU, P, and R denote image-averaged label-set IoU,
		precision, and recall.
		Bold marks the highest value in each column.}
	\label{tab:mlic_nus}
	\small
	\setlength{\tabcolsep}{4pt}
	\begin{tabular}{llccccccccc}
		\toprule
		& & \multicolumn{3}{c}{1-shot}
		& \multicolumn{3}{c}{5-shot}
		& \multicolumn{3}{c}{10-shot} \\
		\cmidrule(lr){3-5}
		\cmidrule(lr){6-8}
		\cmidrule(lr){9-11}
		Encoder & BG
		& IoU & P & R
		& IoU & P & R
		& IoU                         & P              & R              \\
		\midrule
		DINO ViT-S/8 & Yes
		& 14.71 & \textbf{30.84} & 15.40
		& 17.51 & 30.38 & \textbf{31.74}
		& 17.30                       & 30.46          & 30.85          \\
		DINO ViT-S/8 & No
		& 13.20 & 25.20 & 16.88
		& 18.70 & 35.93 & 19.34
		& 19.39                       & 37.22          & 20.00          \\
		\addlinespace
		DINO ViT-S/16 & Yes
		& 14.03 & 29.71 & 14.76
		& 16.97 & 31.26 & 26.25
		& 16.92                       & 29.07          & \textbf{31.77} \\
		DINO ViT-S/16 & No
		& \textbf{14.72} & 26.76 & 19.15
		& \textbf{19.59} & \textbf{36.86} & 20.33
		& \textbf{20.28}              & \textbf{38.42} & 21.03          \\
		\addlinespace
		DINO ViT-B/8 & Yes
		& 13.94 & 29.69 & 15.46
		& 17.08 & 30.13 & 28.69
		& 17.17                       & 30.51          & 28.67          \\
		DINO ViT-B/8 & No
		& 13.14 & 26.44 & 14.49
		& 18.09 & 34.60 & 20.16
		& 18.91                       & 36.04          & 20.65          \\
		\addlinespace
		DINO ViT-B/16 & Yes
		& 13.23 & 26.98 & 18.39
		& 16.49 & 30.50 & 25.87
		& 16.35                       & 30.32          & 25.55          \\
		DINO ViT-B/16 & No
		& 14.17 & 27.97 & 15.25
		& 19.50 & 36.73 & 20.87
		& 20.12                       & 37.95          & 21.44          \\
		\midrule
		MAE ViT-B/16 & Yes
		& 4.47 & 6.82 & 9.38
		& 6.42 & 9.12 & 15.23
		& 7.20                        & 10.01          & 17.37          \\
		MAE ViT-B/16 & No
		& 3.51 & 4.71 & 8.65
		& 8.48 & 10.84 & 23.79
		& 8.51                        & 10.81          & 23.98          \\
		\addlinespace
		MAE ViT-L/16 & Yes
		& 3.15 & 3.19 & \textbf{74.39}
		& 4.36 & 6.37 & 11.70
		& 5.03                        & 6.98           & 14.35          \\
		MAE ViT-L/16 & No
		& 2.88 & 4.45 & 5.17
		& 7.28 & 10.12 & 14.98
		& 4.99                        & 7.80           & 9.98           \\
		\addlinespace
		MAE ViT-H/14 & Yes
		& 3.18 & 3.35 & 39.61
		& 3.87 & 4.93 & 17.20
		& 4.21                        & 5.38           & 16.66          \\
		MAE ViT-H/14 & No
		& 2.60 & 2.64 & 63.07
		& 6.10 & 9.20 & 10.53
		& 4.37                        & 6.49           & 11.51          \\
		\bottomrule
	\end{tabular}
\end{table*}

\begin{table*}[t]
	\centering
	\caption{MLIC results (\%) on VG500. IoU, P, and R denote image-averaged label-set IoU, precision, and recall. Bold marks the highest value in each column.}
	\label{tab:mlic_vg500}
	\small
	\setlength{\tabcolsep}{4pt}
	\begin{tabular}{llccccccccc}
		\toprule
		                                                               &     &           \multicolumn{3}{c}{1-shot}            &           \multicolumn{3}{c}{5-shot}            &           \multicolumn{3}{c}{10-shot}           \\ \cmidrule
		(lr){3-5}
		\cmidrule(lr){6-8}
		\cmidrule(lr){9-11}
		Encoder & BG  &      IoU      &       P        &       R        &      IoU      &       P        &       R        &      IoU      &       P        &       R        \\ \midrule
		DINO ViT-S/8                                                   & Yes &     5.89      & \textbf{34.26} &     11.73      &     6.84      & \textbf{40.27} &     12.60      &     6.76      & \textbf{40.82} &     12.01      \\
		DINO ViT-S/8                                                   & No  &     5.94      &     32.09      &     13.77      &     6.83      &     37.41      &     14.45      &     6.84      &     37.51      &     14.13      \\
		\addlinespace
		DINO ViT-S/16                                  & Yes &     5.61      &     28.94      &     12.57      &     6.73      &     33.34      &     15.27      &     6.76      &     33.84      &     15.29      \\
		DINO ViT-S/16                                                  & No  &     5.77      &     27.51      &     16.30      &     6.55      &     30.83      &     17.72      &     6.69      &     31.47      &     17.36      \\
		\addlinespace
		DINO ViT-B/8                                   & Yes &     6.02      &     32.14      &     17.02      & \textbf{7.05} &     37.15      &     18.61      & \textbf{7.11} &     37.39      &     18.58      \\
		DINO ViT-B/8                                                   & No  & \textbf{6.22} &     31.88      &     18.00      &     7.01      &     34.44      &     22.84      &     7.00      &     34.27      &     22.30      \\
		\addlinespace
		DINO ViT-B/16                                  & Yes &     5.82      &     29.07      &     14.54      &     6.91      &     33.64      &     17.46      &     6.93      &     34.29      &     17.46      \\
		DINO ViT-B/16                                                  & No  &     6.09      &     28.52      &     20.08      &     6.88      &     31.47      &     21.46      &     6.93      &     31.54      &     21.11      \\ \midrule
		MAE ViT-B/16                                                   & Yes &     3.17      &      3.40      &     34.07      &     4.16      &      5.70      &     13.95      &     4.49      &      6.01      &     15.64      \\
		MAE ViT-B/16                                                   & No  &     3.56      &      4.36      &     17.38      &     4.33      &      5.97      &     14.45      &     4.40      &      6.04      &     14.95      \\
		\addlinespace
		MAE ViT-L/16                                   & Yes &     3.00      &      3.18      & \textbf{40.44} &     3.22      &      4.30      &     13.59      &     3.42      &      4.32      &     16.43      \\
		MAE ViT-L/16                                                   & No  &     2.80      &      3.54      &     16.58      &     3.49      &      4.52      &     16.79      &     3.48      &      4.72      &     14.89      \\
		\addlinespace
		MAE ViT-H/14                                   & Yes &     2.79      &      3.03      &     33.18      &     2.93      &      3.39      &     26.02      &     3.02      &      3.64      &     21.15      \\
		MAE ViT-H/14                                                   & No  &     2.68      &      3.04      &     30.53      &     3.20      &      3.69      & \textbf{31.14} &     3.15      &      3.89      & \textbf{23.09} \\ \bottomrule
	\end{tabular}
\end{table*}

\subsection{Object Detection}
\label{sec:supp_od_results}

We compare the AMC-based baseline with an SSP-based detection pipeline. Both use the same single-object training images, frozen recognition encoder, and prototype-matching rule within each matched configuration. The comparison changes region generation, crop masking, and post-processing, and it therefore evaluates complete detection pipelines rather than proposal quality in isolation.

\subsubsection{SSP-Based Detection Pipeline}
\label{sec:supp_ssp_setup}
The SSP-based pipeline uses Selective Search in quality mode to generate candidate boxes. Images are resized while preserving their aspect ratio, with the longer side limited to 400 pixels. The proposals are then mapped back to the original image coordinates. We retain boxes whose area is between 0.5\% and 80\% of the image area and whose aspect ratio is at most 5. Candidates are processed in their original order, with overlap suppression at an IoU threshold of 0.5, retaining at most 100 proposals per image. Both pipelines use the same frozen recognition encoder and class prototypes. SSP encodes rectangular RGB crops, whereas AMC sets pixels outside each candidate mask to black before encoding the crop. Predictions with cosine similarity scores below 0.2 are discarded in both pipelines. SSP then applies score-based, class-wise NMS at an IoU threshold of 0.5, while AMC applies no additional NMS. This final NMS is separate from the class-agnostic proposal filtering performed before recognition.

\subsubsection{Results}
\paragraph{VOC\textsubscript{od}.} 
Table~\ref{tab:od_voc_comparison} shows that AMC outperforms SSP across all evaluated DINO configurations. With ten plain-background images per category, DINO ViT-B/8 achieves 34.39\% Box $\mathrm{AP}_{50}$ with AMC, compared with 8.57\% with SSP. For this AMC configuration, increasing the training set from one to five examples improves AP from 29.16\% to 34.12\%, while ten examples yield only a further 0.27 percentage-point gain. However, SSP consistently performs better with MAE ViT-L/16 and ViT-H/14. The preferred detection pipeline therefore depends on the encoder.

\paragraph{COCO\textsubscript{od}.} Table~\ref{tab:od_coco_comparison} shows a similar encoder-dependent pattern. With ten plain-background images per category, DINO ViT-S/8 achieves 7.05\% Box $\mathrm{AP}_{50:95}$ with AMC, compared with 1.88\% with SSP. Its AMC performance increases from 6.00\% at 1-shot to 6.87\% at 5-shot and 7.05\% at 10-shot, indicating limited gains from additional category examples. Additional examples update category prototypes but leave candidate regions unchanged.

\begin{table*}[t]
	\centering
	\caption{OD results on the VOC split specified in Section~\ref{sec:supp_o2m_datasets}, measured by Box $\mathrm{AP}_{50}$ (\%). Bold marks the higher score between the SSP- and AMC-based pipelines under each matched configuration.}
	\label{tab:od_voc_comparison}
	\small
	\setlength{\tabcolsep}{8pt}
	\begin{tabular}{lllccc}
		\toprule
		Encoder                       & BG  & Method &     1-shot     &     5-shot     &    10-shot     \\ \midrule
		DINO ViT-S/8                  & Yes & SSP    &      7.06      &     10.16      &     10.20      \\
		                              &     & AMC   & \textbf{23.67} & \textbf{30.88} & \textbf{31.15} \\
		\addlinespace
		DINO ViT-S/8  & No  & SSP    &      7.28      &     10.15      &     10.27      \\
		                              &     & AMC   & \textbf{26.72} & \textbf{31.78} & \textbf{32.06} \\ \midrule
		DINO ViT-S/16                 & Yes & SSP    &      5.76      &      9.43      &      9.59      \\
		                              &     & AMC   & \textbf{19.02} & \textbf{26.67} & \textbf{27.25} \\
		\addlinespace
		DINO ViT-S/16 & No  & SSP    &      6.77      &      9.64      &      9.52      \\
		                              &     & AMC   & \textbf{23.28} & \textbf{28.03} & \textbf{28.16} \\ \midrule
		DINO ViT-B/8                  & Yes & SSP    &      6.25      &      9.06      &      9.24      \\
		                              &     & AMC   & \textbf{21.91} & \textbf{30.34} & \textbf{31.90} \\
		\addlinespace
		DINO ViT-B/8  & No  & SSP    &      6.48      &      8.69      &      8.57      \\
		                              &     & AMC   & \textbf{29.16} & \textbf{34.12} & \textbf{34.39} \\ \midrule
		DINO ViT-B/16                 & Yes & SSP    &      6.80      &     10.11      &     10.18      \\
		                              &     & AMC   & \textbf{21.26} & \textbf{28.31} & \textbf{29.28} \\
		\addlinespace
		DINO ViT-B/16 & No  & SSP    &      7.32      &     10.02      &     10.04      \\
		                              &     & AMC   & \textbf{26.52} & \textbf{30.83} & \textbf{31.12} \\ \midrule
		MAE ViT-B/16                  & Yes & SSP    & \textbf{0.80}  &      1.22      &      1.57      \\
		                              &     & AMC   &      0.77      & \textbf{3.27}  & \textbf{2.57}  \\
		\addlinespace
		MAE ViT-B/16  & No  & SSP    &      2.15      &      4.12      &      3.68      \\
		                              &     & AMC   & \textbf{3.82}  & \textbf{5.40}  & \textbf{5.83}  \\ \midrule
		MAE ViT-L/16                  & Yes & SSP    & \textbf{0.76}  & \textbf{1.64}  & \textbf{1.80}  \\
		                              &     & AMC   &      0.52      &      0.55      &      0.99      \\
		\addlinespace
		MAE ViT-L/16  & No  & SSP    & \textbf{1.26}  & \textbf{2.77}  & \textbf{2.95}  \\
		                              &     & AMC   &      0.49      &      1.15      &      1.09      \\ \midrule
		MAE ViT-H/14                  & Yes & SSP    & \textbf{0.53}  & \textbf{0.63}  & \textbf{0.68}  \\
		                              &     & AMC   &      0.23      &      0.19      &      0.23      \\
		\addlinespace
		MAE ViT-H/14  & No  & SSP    & \textbf{0.50}  & \textbf{1.55}  & \textbf{1.60}  \\
		                              &     & AMC   &      0.31      &      0.62      &      0.71      \\ \bottomrule
	\end{tabular}
\end{table*}

\begin{table*}[t]
	\centering
	\caption{OD results on the local COCO test partition, measured by Box $\mathrm{AP}_{50:95}$ (\%). Bold marks the higher score between the SSP- and AMC-based pipelines under each matched configuration.}
	\label{tab:od_coco_comparison}
	\small
	\setlength{\tabcolsep}{8pt}
	\begin{tabular}{lllccc}
		\toprule
		Encoder                       & BG  & Method &    1-shot     &    5-shot     &    10-shot    \\ \midrule
		DINO ViT-S/8                  & Yes & SSP    &     1.18      &     1.48      &     1.55      \\
		                              &     & Mask   & \textbf{5.50} & \textbf{6.21} & \textbf{6.49} \\
		\addlinespace
		DINO ViT-S/8  & No  & SSP    &     1.57      &     1.84      &     1.88      \\
		                              &     & Mask   & \textbf{6.00} & \textbf{6.87} & \textbf{7.05} \\ \midrule
		DINO ViT-S/16                 & Yes & SSP    &     1.01      &     1.25      &     1.30      \\
		                              &     & Mask   & \textbf{4.15} & \textbf{4.51} & \textbf{4.62} \\
		\addlinespace
		DINO ViT-S/16 & No  & SSP    &     1.36      &     1.64      &     1.69      \\
		                              &     & Mask   & \textbf{4.60} & \textbf{5.27} & \textbf{5.44} \\ \midrule
		DINO ViT-B/8                  & Yes & SSP    &     1.13      &     1.41      &     1.43      \\
		                              &     & Mask   & \textbf{5.18} & \textbf{5.94} & \textbf{6.10} \\
		\addlinespace
		DINO ViT-B/8  & No  & SSP    &     1.40      &     1.62      &     1.65      \\
		                              &     & Mask   & \textbf{6.02} & \textbf{6.76} & \textbf{6.97} \\ \midrule
		DINO ViT-B/16                 & Yes & SSP    &     1.12      &     1.42      &     1.47      \\
		                              &     & Mask   & \textbf{4.31} & \textbf{4.87} & \textbf{4.98} \\
		\addlinespace
		DINO ViT-B/16 & No  & SSP    &     1.57      &     1.79      &     1.86      \\
		                              &     & Mask   & \textbf{5.00} & \textbf{5.70} & \textbf{5.86} \\ \midrule
		MAE ViT-B/16                  & Yes & SSP    &     0.14      &     0.15      &     0.18      \\
		                              &     & Mask   & \textbf{0.32} & \textbf{0.18} & \textbf{0.32} \\
		\addlinespace
		MAE ViT-B/16  & No  & SSP    &     0.61      &     0.63      &     0.60      \\
		                              &     & Mask   & \textbf{0.73} & \textbf{0.92} & \textbf{0.80} \\ \midrule
		MAE ViT-L/16                  & Yes & SSP    & \textbf{0.12} & \textbf{0.12} & \textbf{0.17} \\
		                              &     & Mask   &     0.03      &     0.05      &     0.07      \\
		\addlinespace
		MAE ViT-L/16  & No  & SSP    & \textbf{0.43} & \textbf{0.45} & \textbf{0.45} \\
		                              &     & Mask   &     0.15      &     0.16      &     0.16      \\ \midrule
		MAE ViT-H/14                  & Yes & SSP    & \textbf{0.08} & \textbf{0.07} & \textbf{0.11} \\
		                              &     & Mask   &     0.01      &     0.03      &     0.06      \\
		\addlinespace
		MAE ViT-H/14  & No  & SSP    & \textbf{0.23} & \textbf{0.28} & \textbf{0.27} \\
		                              &     & Mask   &     0.07      &     0.09      &     0.08      \\ \bottomrule
	\end{tabular}
\end{table*}

\subsection{Semantic Segmentation} \label{sec:supp_ss_results} 
Tables~\ref{tab:ss_voc} and~\ref{tab:ss_coco} report mIoU, pixel accuracy (PA), and mean class accuracy (mAcc). Metric definitions and the treatment of background and ignored pixels are provided in Section~\ref{sec:supp_o2m_protocol}.
\paragraph{VOC\textsubscript{ss}.} Table~\ref{tab:ss_voc} shows that single-object category supervision can support pixel-level semantic labeling when combined with class-agnostic masks. With ten plain-background images per category, DINO ViT-B/8 achieves the highest mIoU of 46.52\% and mAcc of 66.01\%. DINO ViT-S/8 instead achieves the highest PA of 82.54\%, but a lower mIoU of 45.27\%. The different rankings show that greater overall pixel accuracy does not necessarily imply better class-averaged segmentation quality. Reporting mIoU alongside PA therefore provides a more complete assessment of the predictions. 
\paragraph{COCO\textsubscript{ss}.} 
Table~\ref{tab:ss_coco} shows that DINO ViT-S/8 with plain-background training images achieves the highest mIoU at all three training set sizes, reaching 24.43\% at 10-shot. Under matched background and sample-count conditions, the DINO /8 variants consistently outperform their /16 counterparts in mIoU. This pattern is consistent with the potential benefit of finer spatial features, although changing the encoder also changes the features used for category recognition.

\begin{table*}[t]
	\centering
	\caption{SS results (\%) on VOC.
		PA and mAcc denote pixel accuracy and mean class accuracy.
		Background is included, and ignored pixels are excluded.
		Bold marks the highest value in each column.}
	\label{tab:ss_voc}
	\small
	\setlength{\tabcolsep}{4pt}
	\begin{tabular}{llccccccccc}
		\toprule
		                                                               &     &            \multicolumn{3}{c}{1-shot}            &            \multicolumn{3}{c}{5-shot}            &           \multicolumn{3}{c}{10-shot}            \\ \cmidrule
		(lr){3-5}
		\cmidrule(lr){6-8}
		\cmidrule(lr){9-11}
		Encoder & BG  &      mIoU      &       PA       &      mAcc      &      mIoU      &       PA       &      mAcc      &      mIoU      &       PA       &      mAcc      \\ \midrule
		DINO ViT-S/8                                                   & Yes &     36.61      &     78.29      &     49.04      & \textbf{46.44} &     81.38      &     61.02      &     46.41      &     80.57      &     61.62      \\
		DINO ViT-S/8                                                   & No  & \textbf{41.59} & \textbf{81.32} &     57.97      &     43.07      & \textbf{81.77} &     60.47      &     45.27      & \textbf{82.54} &     62.51      \\
		\addlinespace
		DINO ViT-S/16                                  & Yes &     35.03      &     74.07      &     50.07      &     41.76      &     77.41      &     64.12      &     42.50      &     76.56      &     64.00      \\
		DINO ViT-S/16                                                  & No  &     37.30      &     76.68      &     58.18      &     40.57      &     78.10      &     63.30      &     41.91      &     78.03      &     65.43      \\
		\addlinespace
		DINO ViT-B/8                                   & Yes &     30.29      &     75.31      &     41.83      &     44.26      &     79.17      &     60.00      &     42.95      &     77.41      &     59.40      \\
		DINO ViT-B/8                                                   & No  &     38.20      &     77.49      &     56.37      &     45.16      &     79.46      & \textbf{64.49} & \textbf{46.52} &     79.53      & \textbf{66.01} \\
		\addlinespace
		DINO ViT-B/16                                  & Yes &     29.80      &     70.04      &     45.93      &     39.04      &     72.01      &     62.20      &     37.21      &     70.09      &     61.73      \\
		DINO ViT-B/16                                                  & No  &     39.09      &     72.04      & \textbf{61.43} &     37.73      &     71.07      &     62.56      &     39.34      &     70.81      &     63.54      \\ \midrule
		MAE ViT-B/16                                                   & Yes &      5.08      &     56.92      &     10.05      &     15.95      &     62.28      &     25.22      &     12.38      &     60.04      &     22.70      \\
		MAE ViT-B/16                                                   & No  &      9.24      &     59.53      &     18.45      &     13.93      &     62.31      &     25.84      &     13.79      &     62.66      &     24.27      \\
		\addlinespace
		MAE ViT-L/16                                   & Yes &      4.49      &     35.32      &     10.71      &      4.58      &     35.14      &     12.04      &      7.09      &     36.24      &     17.23      \\
		MAE ViT-L/16                                                   & No  &      3.00      &     33.49      &      8.07      &      6.92      &     35.80      &     12.96      &      5.23      &     35.05      &     10.63      \\
		\addlinespace
		MAE ViT-H/14                                   & Yes &      3.76      &     48.55      &      7.76      &      4.34      &     48.38      &      6.78      &      4.80      &     49.08      &     10.55      \\
		MAE ViT-H/14                                                   & No  &      3.54      &     48.32      &      8.50      &      3.77      &     48.37      &      6.95      &      4.19      &     48.82      &      8.37      \\ \bottomrule
	\end{tabular}
\end{table*}

\begin{table*}[t]
	\centering
	\caption{SS results (\%) on COCO using instance-derived semantic labels.
		PA and mAcc denote pixel accuracy and mean class accuracy.
		Background is included, and ignored pixels are excluded.
		Bold marks the highest value in each column.}
	\label{tab:ss_coco}
	\small
	\setlength{\tabcolsep}{4pt}
	\begin{tabular}{llccccccccc}
		\toprule
		                                                               &     &            \multicolumn{3}{c}{1-shot}            &            \multicolumn{3}{c}{5-shot}            &           \multicolumn{3}{c}{10-shot}            \\ \cmidrule
		(lr){3-5}
		\cmidrule(lr){6-8}
		\cmidrule(lr){9-11}
		Encoder & BG  &      mIoU      &       PA       &      mAcc      &      mIoU      &       PA       &      mAcc      &      mIoU      &       PA       &      mAcc      \\ \midrule
		DINO ViT-S/8                                                   & Yes &     20.00      &     69.96      &     37.09      &     22.65      &     69.93      &     42.93      &     23.53      &     70.64      &     44.35      \\
		DINO ViT-S/8                                                   & No  & \textbf{20.42} & \textbf{71.72} &     39.63      & \textbf{23.30} & \textbf{73.52} &     44.87      & \textbf{24.43} & \textbf{73.98} &     45.94      \\
		\addlinespace
		DINO ViT-S/16                                  & Yes &     15.99      &     63.30      &     34.04      &     18.15      &     63.38      &     39.65      &     19.11      &     64.51      &     41.06      \\
		DINO ViT-S/16                                                  & No  &     17.99      &     67.21      &     35.91      &     20.24      &     68.81      &     41.63      &     21.06      &     69.30      &     43.01      \\
		\addlinespace
		DINO ViT-B/8                                   & Yes &     19.22      &     67.58      &     36.21      &     21.07      &     67.04      &     41.95      &     22.14      &     67.60      &     43.60      \\
		DINO ViT-B/8                                                   & No  &     19.98      &     68.79      & \textbf{39.85} &     21.94      &     70.27      &     44.76      &     23.31      &     71.18      &     45.81      \\
		\addlinespace
		DINO ViT-B/16                                  & Yes &     16.53      &     59.80      &     35.61      &     18.76      &     60.26      &     42.76      &     20.01      &     60.99      &     44.49      \\
		DINO ViT-B/16                                                  & No  &     18.91      &     63.99      &     39.56      &     20.64      &     65.39      & \textbf{45.00} &     21.81      &     65.74      & \textbf{46.15} \\ \midrule
		MAE ViT-B/16                                                   & Yes &      2.59      &     54.04      &      5.50      &      2.43      &     54.31      &      6.98      &      3.56      &     54.90      &      8.46      \\
		MAE ViT-B/16                                                   & No  &      4.12      &     55.26      &      9.15      &      5.52      &     57.68      &     10.87      &      5.06      &     58.13      &     10.64      \\
		\addlinespace
		MAE ViT-L/16                                   & Yes &      1.19      &     32.58      &      3.83      &      1.64      &     32.61      &      4.69      &      2.39      &     33.26      &      7.75      \\
		MAE ViT-L/16                                                   & No  &      1.86      &     32.76      &      4.05      &      2.62      &     33.56      &      6.96      &      2.43      &     33.55      &      6.54      \\
		\addlinespace
		MAE ViT-H/14                                   & Yes &      0.98      &     46.94      &      2.70      &      0.99      &     46.92      &      2.63      &      1.50      &     47.34      &      4.34      \\
		MAE ViT-H/14                                                   & No  &      1.28      &     46.95      &      2.96      &      1.57      &     47.37      &      3.16      &      1.33      &     47.61      &      4.09      \\ \bottomrule
	\end{tabular}
\end{table*}

\subsection{Instance Segmentation} \label{sec:supp_is_results} 
Tables~\ref{tab:is_voc} and~\ref{tab:is_coco} report Mask $\mathrm{AP}_{50:95}$ as the primary metric, alongside $\mathrm{AP}_{50}$ and $\mathrm{AP}_{75}$. The latter two metrics assess instance predictions under different mask-overlap requirements. Evaluation details are provided in Section~\ref{sec:supp_o2m_protocol}. 

\paragraph{VOC\textsubscript{is}.} Table~\ref{tab:is_voc} shows that DINO ViT-S/8 achieves the highest Mask $\mathrm{AP}_{50:95}$ at each training set size. Its best result is 21.00\%, obtained with five contextual-background images per category. For this configuration, $\mathrm{AP}_{50}$ reaches 35.52\%, whereas $\mathrm{AP}_{75}$ falls to 18.75\%. The gap indicates that many predictions accepted under the looser overlap criterion do not satisfy the stricter requirement. Accurate instance delineation therefore remains a challenge, even when category supervision is sufficient to produce useful predictions.

\paragraph{COCO\textsubscript{is}.} Table~\ref{tab:is_coco} shows that DINO ViT-S/8 with ten plain-background images per category achieves the highest Mask $\mathrm{AP}_{50:95}$ of 6.39\%. Under the same training conditions, ViT-B/8 obtains a slightly higher $\mathrm{AP}_{50}$ of 11.42\% versus 11.36\%, but lower $\mathrm{AP}_{75}$ and $\mathrm{AP}_{50:95}$. Thus, the ranking under a loose matching criterion does not necessarily hold when more precise mask overlap is required. The low overall AP highlights the limitations of the current combination of class-agnostic masks and prototype-based recognition for instance-level prediction.

\begin{table*}[t]
	\centering
	\caption{IS results (\%) on VOC using COCO-style evaluation
		with void pixels excluded.
		All AP metrics use masks, and AP denotes
		$\mathrm{AP}_{50:95}$.
		Bold marks the highest value in each column.}
	\label{tab:is_voc}
	\small
	\setlength{\tabcolsep}{4pt}
	\begin{tabular}{llccccccccc}
		\toprule
		                                                               &     &            \multicolumn{3}{c}{1-shot}            &            \multicolumn{3}{c}{5-shot}            &           \multicolumn{3}{c}{10-shot}            \\ \cmidrule
		(lr){3-5}
		\cmidrule(lr){6-8}
		\cmidrule(lr){9-11}
		Encoder & BG  &       AP       &   AP$_{50}$    &   AP$_{75}$    &       AP       &   AP$_{50}$    &   AP$_{75}$    &       AP       &   AP$_{50}$    &   AP$_{75}$    \\ \midrule
		DINO ViT-S/8                                                   & Yes &     16.96      &     28.28      &     14.84      & \textbf{21.00} & \textbf{35.52} & \textbf{18.75} & \textbf{20.89} & \textbf{35.63} & \textbf{18.60} \\
		DINO ViT-S/8                                                   & No  & \textbf{18.58} & \textbf{30.21} & \textbf{16.42} &     20.38      &     33.62      &     18.30      &     20.69      &     34.71      &     18.44      \\
		\addlinespace
		DINO ViT-S/16                                  & Yes &     12.84      &     20.13      &     12.63      &     15.89      &     26.11      &     15.28      &     16.48      &     26.81      &     16.13      \\
		DINO ViT-S/16                                                  & No  &     13.74      &     22.52      &     13.19      &     16.28      &     26.01      &     15.70      &     16.57      &     26.98      &     15.99      \\
		\addlinespace
		DINO ViT-B/8                                   & Yes &     14.38      &     25.90      &     12.45      &     17.94      &     32.40      &     15.80      &     18.18      &     32.10      &     16.14      \\
		DINO ViT-B/8                                                   & No  &     16.73      &     29.30      &     14.74      &     18.13      &     32.52      &     15.67      &     19.00      &     34.05      &     16.68      \\
		\addlinespace
		DINO ViT-B/16                                  & Yes &     12.90      &     21.30      &     12.32      &     15.60      &     25.76      &     15.11      &     15.57      &     25.66      &     15.20      \\
		DINO ViT-B/16                                                  & No  &     15.66      &     25.74      &     15.18      &     16.97      &     27.09      &     16.29      &     16.83      &     27.72      &     16.08      \\ \midrule
		MAE ViT-B/16                                                   & Yes &      0.34      &      0.55      &      0.31      &      3.37      &      6.09      &      3.69      &      2.03      &      3.91      &      1.89      \\
		MAE ViT-B/16                                                   & No  &      2.34      &      4.81      &      1.61      &      4.02      &      6.99      &      3.42      &      3.73      &      6.28      &      3.80      \\
		\addlinespace
		MAE ViT-L/16                                   & Yes &      0.35      &      0.74      &      0.07      &      0.08      &      0.33      &      0.03      &      0.28      &      0.98      &      0.07      \\
		MAE ViT-L/16                                                   & No  &      0.00      &      0.04      &      0.00      &      0.22      &      0.54      &      0.30      &      0.21      &      0.39      &      0.30      \\
		\addlinespace
		MAE ViT-H/14                                   & Yes &      0.04      &      0.12      &      0.00      &      0.11      &      0.33      &      0.01      &      0.04      &      0.18      &      0.00      \\
		MAE ViT-H/14                                                   & No  &      0.06      &      0.16      &      0.05      &      0.10      &      0.18      &      0.08      &      0.16      &      0.35      &      0.15      \\ \bottomrule
	\end{tabular}
\end{table*}

\begin{table*}[t]
	\centering
	\caption{IS results (\%) on COCO. All AP metrics use masks, and AP denotes $\mathrm{AP}_{50:95}$. Bold marks the highest value in each column.}
	\label{tab:92188b54}
	\label{tab:is_coco}
	\small
	\setlength{\tabcolsep}{4pt}
	\begin{tabular}{llccccccccc}
		\toprule
		                                                               &     &          \multicolumn{3}{c}{1-shot}           &           \multicolumn{3}{c}{5-shot}           &          \multicolumn{3}{c}{10-shot}           \\ \cmidrule
		(lr){3-5}
		\cmidrule(lr){6-8}
		\cmidrule(lr){9-11}
		Encoder & BG  &      AP       &   AP$_{50}$   &   AP$_{75}$   &      AP       &   AP$_{50}$    &   AP$_{75}$   &      AP       &   AP$_{50}$    &   AP$_{75}$   \\ \midrule
		DINO ViT-S/8                                                   & Yes &     4.99      &     8.97      &     4.90      &     5.73      &     10.26      &     5.63      &     6.00      &     10.72      &     5.88      \\
		DINO ViT-S/8                                                   & No  & \textbf{5.40} &     9.66      & \textbf{5.33} & \textbf{6.23} &     11.07      & \textbf{6.16} & \textbf{6.39} &     11.36      & \textbf{6.29} \\
		\addlinespace
		DINO ViT-S/16                                  & Yes &     3.76      &     6.91      &     3.69      &     4.08      &      7.54      &     4.00      &     4.21      &      7.83      &     4.10      \\
		DINO ViT-S/16                                                  & No  &     4.17      &     7.53      &     4.08      &     4.80      &      8.68      &     4.75      &     4.95      &      9.00      &     4.86      \\
		\addlinespace
		DINO ViT-B/8                                   & Yes &     4.60      &     8.72      &     4.38      &     5.33      &     10.03      &     5.12      &     5.49      &     10.33      &     5.28      \\
		DINO ViT-B/8                                                   & No  &     5.34      & \textbf{9.95} &     5.12      &     5.97      & \textbf{11.09} &     5.77      &     6.15      & \textbf{11.42} &     5.93      \\
		\addlinespace
		DINO ViT-B/16                                  & Yes &     3.92      &     7.14      &     3.92      &     4.40      &      8.05      &     4.39      &     4.53      &      8.31      &     4.51      \\
		DINO ViT-B/16                                                  & No  &     4.57      &     8.19      &     4.58      &     5.21      &      9.37      &     5.24      &     5.37      &      9.72      &     5.35      \\ \midrule
		MAE ViT-B/16                                                   & Yes &     0.28      &     0.54      &     0.26      &     0.15      &      0.32      &     0.13      &     0.23      &      0.51      &     0.18      \\
		MAE ViT-B/16                                                   & No  &     0.62      &     1.18      &     0.60      &     0.77      &      1.49      &     0.73      &     0.69      &      1.34      &     0.65      \\
		\addlinespace
		MAE ViT-L/16                                   & Yes &     0.01      &     0.03      &     0.00      &     0.01      &      0.03      &     0.00      &     0.02      &      0.07      &     0.01      \\
		MAE ViT-L/16                                                   & No  &     0.05      &     0.15      &     0.03      &     0.06      &      0.17      &     0.02      &     0.06      &      0.18      &     0.03      \\
		\addlinespace
		MAE ViT-H/14                                   & Yes &     0.01      &     0.02      &     0.00      &     0.02      &      0.04      &     0.02      &     0.05      &      0.09      &     0.04      \\
		MAE ViT-H/14                                                   & No  &     0.06      &     0.11      &     0.04      &     0.06      &      0.14      &     0.05      &     0.06      &      0.13      &     0.05      \\ \bottomrule
	\end{tabular}
\end{table*}

\section{Additional Analysis}
\label{sec:supp_additional_analysis}

This section provides the aggregation procedures and numerical summaries underlying Figures~\ref{fig:sample_gains}, \ref{fig:background_effect}, and \ref{fig:encoder_comparison}. All comparisons use the primary metric of each task. OD results include only the AMC-based baseline unless otherwise stated. Interpretations are provided in Section~\ref{sec:discussion} of the main paper.

\subsection{Training Set Size}
\label{sec:pb0d23b7b26eabc90dd423a257d274911}

Table~\ref{tab:feeb89bd} summarizes performance with 1, 5, and 10 training images per category. Within each task group, we average scores equally across the seven encoders, two background conditions, and corresponding datasets. MLIC therefore includes 56 configurations per training set size, while SS and IS each include 28. OD is reported separately for VOC and COCO because their primary metrics differ, with 14 configurations per dataset. The averages are configuration-level summaries, not image-count-weighted averages across datasets. Training subsets are nested within each category and background condition. The encoder and class-agnostic candidate regions remain unchanged across training set sizes. Category prototypes are recomputed from each subset, and the MLIC top-$p$ threshold is selected separately on the validation set for each configuration. Gains are calculated from unrounded average scores and reported in percentage points.

\begin{table*}[t]
	\centering
	\caption{
		Mean performance across training set sizes.
		Scores are percentages, and gains are percentage points
		calculated before rounding.
		OD includes only the AMC-based baseline.
	}
	\label{tab:feeb89bd}
	\small
	\begin{tabular}{@{}llrrrrr@{}}
		\toprule
		Task     & Metric                     & 1-shot & 5-shot & 10-shot & $1\rightarrow5$ & $5\rightarrow10$ \\ \midrule
		MLIC     & Label-set IoU              &  17.52 &  21.75 &   22.29 &           +4.23 &            +0.54 \\
		OD--VOC  & Box $\mathrm{AP}_{50}$     &  14.12 &  18.01 &   18.34 &           +3.89 &            +0.33 \\
		OD--COCO & Box $\mathrm{AP}_{50:95}$  &   3.01 &   3.40 &    3.50 &           +0.39 &            +0.10 \\
		SS       & mIoU                       &  17.07 &  20.32 &   20.76 &           +3.25 &            +0.44 \\
		IS       & Mask $\mathrm{AP}_{50:95}$ &   5.81 &   6.89 &    6.96 &           +1.08 &            +0.07 \\ \bottomrule
	\end{tabular}
\end{table*}

\subsection{Training-Image Backgrounds}
\label{sec:p60ce6f6b5a452a3e3ad72753c36ff5e3}

Table~\ref{tab:d095e89f} compares contextual and plain backgrounds under matched task, dataset, encoder, and training set size. Each pair contains one configuration from each background condition. MLIC contributes $84$ pairs, and each of OD, SS, and IS contributes $42$ pairs, giving 210 pairs in total. A condition wins when it achieves a higher unrounded primary score than its paired counterpart. No ties occur in these comparisons. The overall win rate pools all 210 pairs. Plain backgrounds win in 162 pairs (77.14\%). These rates describe how often a condition performs better, not the size of the improvement. Because the two training collections are generated independently, this comparison does not isolate background content from differences in object appearance.

\begin{table}[t]
	\centering
	\caption{
		Background-condition win counts and rates.
		Comparisons match the task, dataset, encoder,
		and training set size.
		OD uses AMC.
	}
	\label{tab:d095e89f}
	\small
	\begin{tabular}{@{}lrrr@{}}
		\toprule
		Task     & Pairs & Contextual wins &    Plain wins \\ \midrule
		MLIC     &    84 &    30 (35.71\%) &  54 (64.29\%) \\
		OD (AMC) &    42 &      1 (2.38\%) &  41 (97.62\%) \\
		SS       &    42 &    12 (28.57\%) &  30 (71.43\%) \\
		IS       &    42 &     5 (11.90\%) &  37 (88.10\%) \\ \midrule
		Overall  &   210 &    48 (22.86\%) & 162 (77.14\%) \\ \bottomrule
	\end{tabular}
\end{table}

\subsection{Visual Encoders}
\label{sec:pe2a8dbcb92b5678767bb70a2abe71d44}

Table~\ref{tab:4ef62a91} reports pairwise win rates among the four DINO encoders. Each encoder is compared against the other three under matched task, dataset, background condition, and training set size. A win is determined by the task's unrounded primary score. For each encoder, MLIC contributes $72$ comparisons. Each of OD, SS, and IS contributes $36$ comparisons. The overall win rate is computed across all 180 comparisons per encoder, rather than by equally averaging the four task-level rates. ViT-S/8 has the highest overall rate of 79.44\%, while ViT-B/8 has the highest MLIC rate of 73.61\%. For dense prediction tasks, changing the encoder affects both localization and recognition features, so these comparisons assess the complete baseline rather than either component in isolation.

\begin{table}[t]
	\centering
	\caption{
		Pairwise win rates (\%) among DINO encoders.
		Comparisons match the task, dataset, background condition,
		and training set size.
		Overall rates pool all comparisons.
		OD uses AMC.
		Bold marks the highest rate in each column.
	}
	\label{tab:4ef62a91}
	\small
	\begin{tabular}{@{}lrrrrr@{}}
		\toprule
		Encoder  &        Overall &           MLIC &             OD &             SS &              IS \\ \midrule
		ViT-S/8  & \textbf{79.44} &          58.33 & \textbf{86.11} & \textbf{94.44} & \textbf{100.00} \\
		ViT-B/8  &          72.22 & \textbf{73.61} &          80.56 &          66.67 &           66.67 \\
		ViT-B/16 &          35.00 &          45.83 &          33.33 &          22.22 &           27.78 \\
		ViT-S/16 &          13.33 &          22.22 &           0.00 &          16.67 &            5.56 \\ \bottomrule
	\end{tabular}
\end{table}

\subsection{Qualitative Results on VOC} \label{sec:supp_voc_qualitative} 
\begin{figure*}[t]
	\centering
	\includegraphics[
	width=0.95\textwidth,
	height=0.78\textheight,
	keepaspectratio
	]{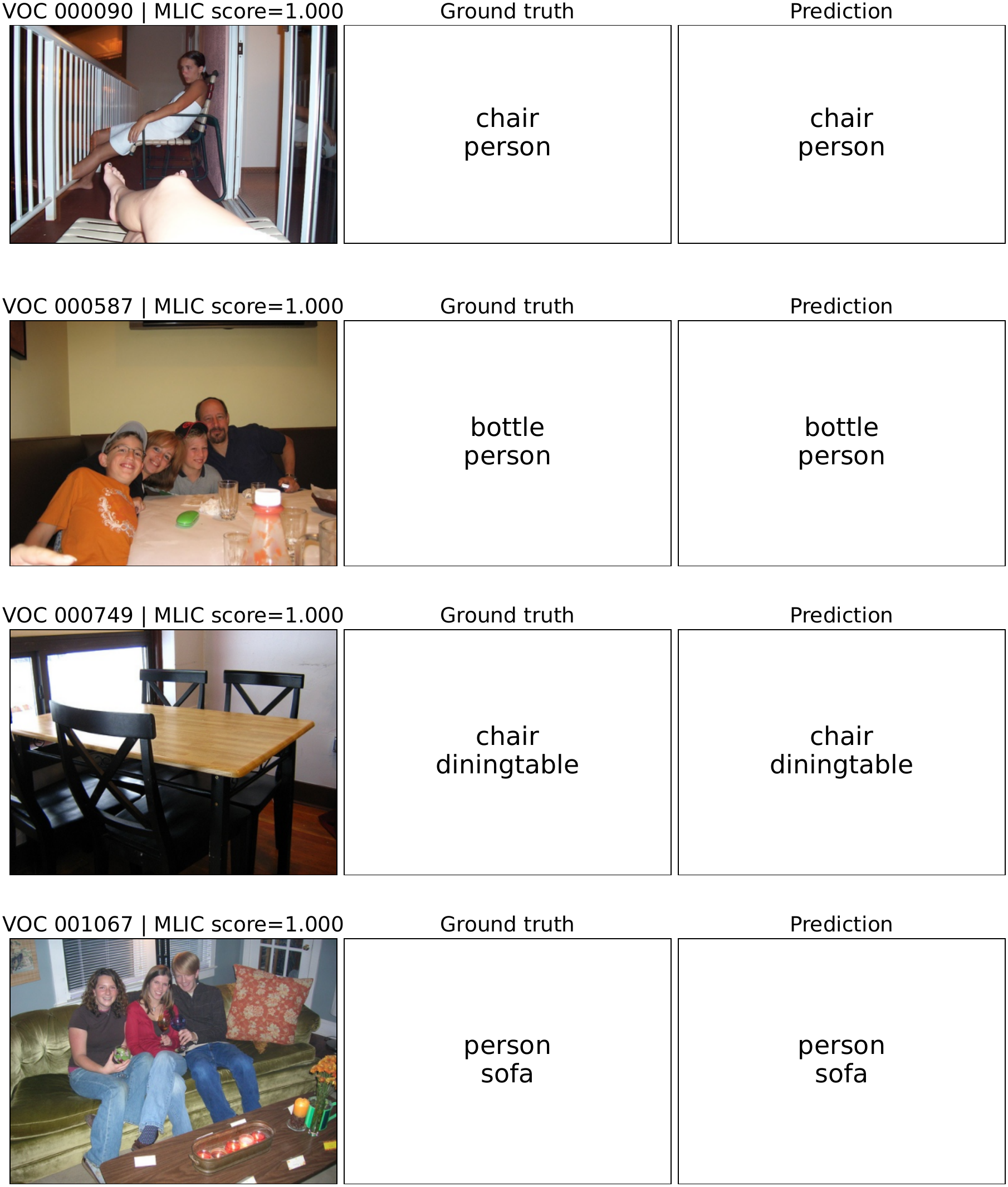}
	\caption{
		Selected MLIC predictions on VOC.
		Each row shows the input, ground truth, and prediction.
		Scores denote per-image label-set IoU.
		The top-$p$ threshold is fixed at the
		validation-selected value of 0.6.
	}
	\label{fig:supp_voc_mlic_cases}
\end{figure*}

\begin{figure*}[t]
	\centering
	\includegraphics[
	width=0.95\textwidth,
	height=0.78\textheight,
	keepaspectratio
	]{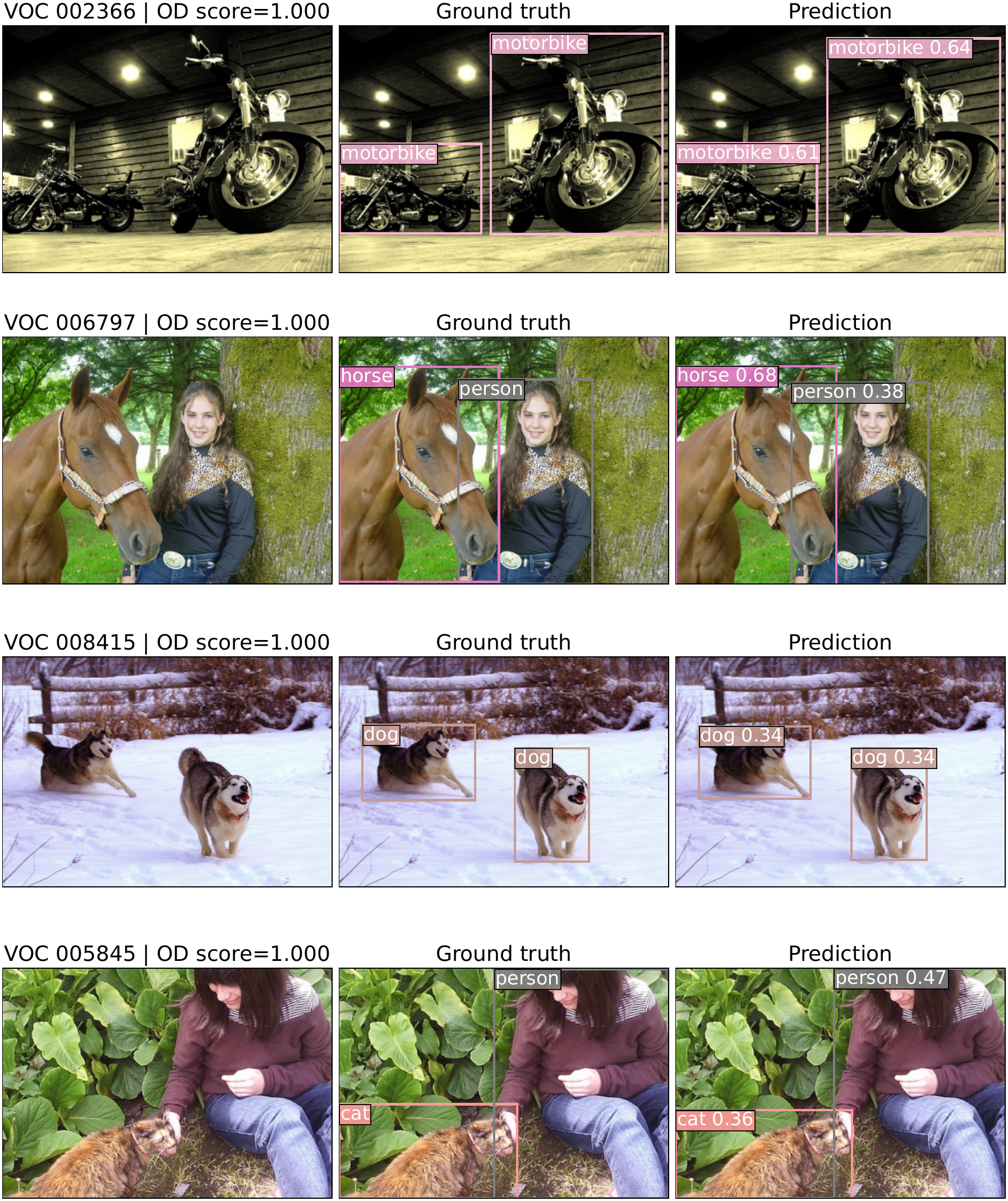}
	\caption{
		Selected OD predictions on the benchmark's VOC test split.
		Each row shows the input, ground truth, and prediction.
		All detections passing the fixed score threshold are shown.
		Case scores denote class-aware matching F1
		at box IoU 0.5, not dataset-level AP.
	}
	\label{fig:supp_voc_od_cases}
\end{figure*}

\begin{figure*}[t]
	\centering
	\includegraphics[
	width=0.95\textwidth,
	height=0.78\textheight,
	keepaspectratio
	]{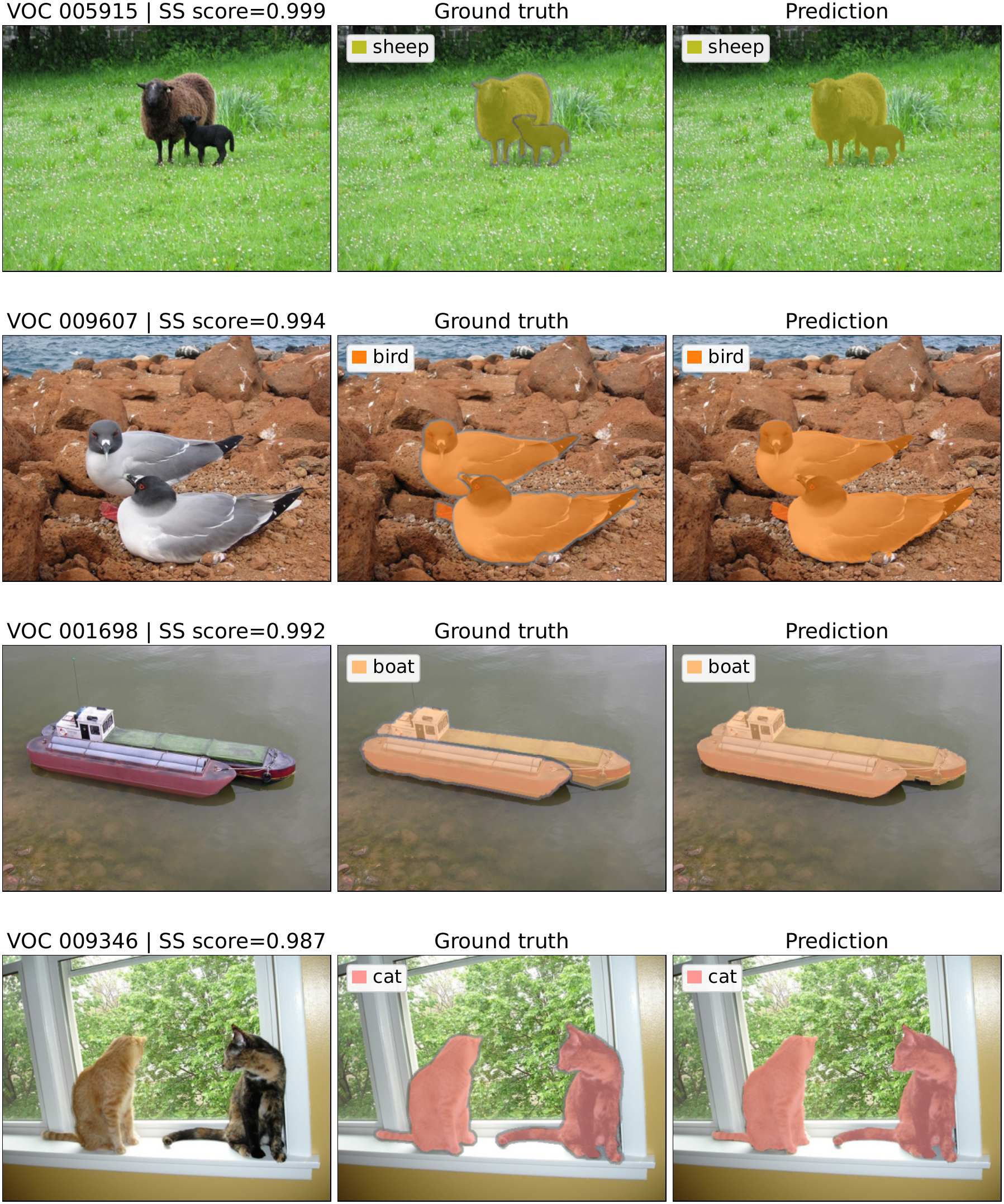}
	\caption{
		Selected SS predictions on VOC.
		Each row shows the input, ground truth, and prediction.
		Colors indicate semantic categories.
		Case scores denote per-image foreground macro IoU
		over categories present in either ground truth or prediction,
		excluding ignored pixels.
		These scores differ from the dataset-level mIoU.
	}
	\label{fig:supp_voc_ss_cases}
\end{figure*}

\begin{figure*}[t]
	\centering
	\includegraphics[
	width=0.95\textwidth,
	height=0.78\textheight,
	keepaspectratio
	]{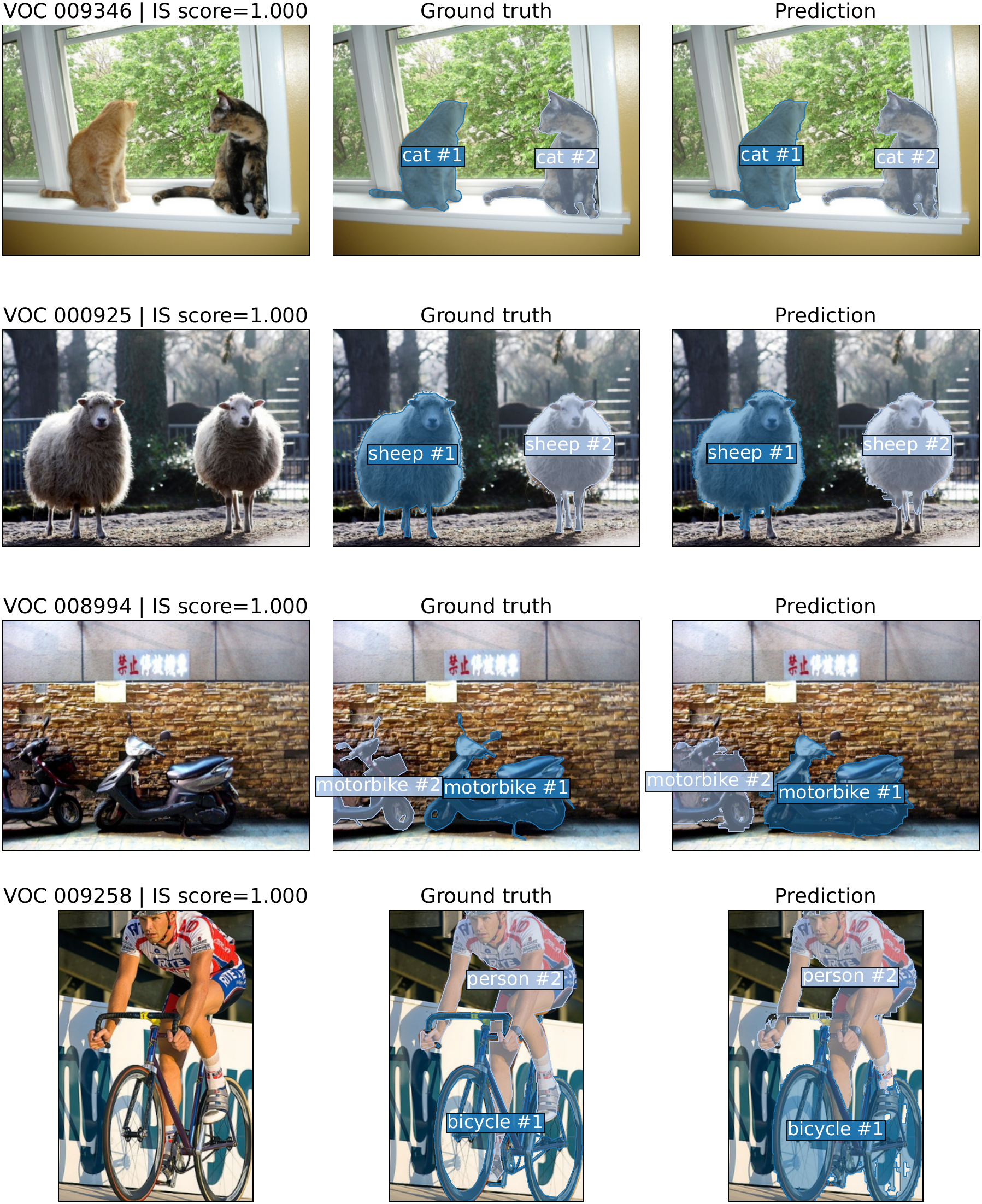}
	\caption{
		Selected IS predictions on VOC.
		Each row shows the input, ground truth, and prediction.
		Colors and identifiers distinguish instances
		and align matched ground-truth and predicted objects.
		Case scores denote class-aware matching F1
		at mask IoU 0.5, not dataset-level AP.
	}
	\label{fig:supp_voc_is_cases}
\end{figure*}

Figures~\ref{fig:supp_voc_mlic_cases}--\ref{fig:supp_voc_is_cases} provide four selected test examples per task, complementing Figure~\ref{fig:qualitative_results}. All examples use DINO ViT-S/8 with ten plain-background training images per category. Each row presents the input, ground truth, and prediction. Predictions are shown without manual correction or additional filtering beyond the task-specific protocol. The examples illustrate the baseline's capabilities rather than its average test performance. Examples are selected after evaluation based on task-specific prediction quality and category diversity. Each image contains at least two ground-truth categories for MLIC or two instances for OD, SS, and IS. Selection does not alter the model, thresholds, or predictions.

\subsection{Code and Data Availability}
\label{sec:supp_code_data}

The benchmark construction scripts, evaluation protocols,
and baseline implementation are publicly available at
\url{https://github.com/t1anhen9/one2multi-learning}.
The synthetic single-object training dataset is available at
\url{https://huggingface.co/datasets/hengtian/one2multi-training}.
It contains 13,620 images with category labels, generation
metadata, and sample manifests for the 1-, 5-, and 10-shot
settings.
For existing datasets, we provide preparation instructions
and split definitions rather than redistribute the original
images and annotations.
These resources support reproducibility and future research
on one2multi learning.
\end{document}